%% file: main.tex
\documentclass[10pt,twocolumn,letterpaper]{article}

\usepackage[pagenumbers]{cvpr}

\input{preamble}

\definecolor{cvprblue}{rgb}{0.21,0.49,0.74}
\usepackage[pagebackref,breaklinks,colorlinks,allcolors=cvprblue]{hyperref}
\usepackage{bm}

\title{Astrea: A MOE-based Visual Understanding Model with Progressive Alignment}

\author{\textbf{Xiaoda Yang}$^1*$\qquad 
\textbf{JunYu Lu}$^2*$\qquad 
\textbf{Hongshun Qiu}$^3$\qquad 
\textbf{Sijing Li}$^1$\qquad 
\textbf{Hao Li}$^4$\qquad \\
\textbf{Shengpeng Ji}$^1$\qquad 
\textbf{Xudong Tang}$^5$\qquad 
\textbf{Jiayang Xu}$^1$\qquad 
\textbf{Jiaqi Duan}$^6$\qquad 
\textbf{Ziyue Jiang}$^1$\qquad\\ 
\textbf{Cong Lin}$^2$\qquad 
\textbf{Sihang Cai}$^1$\qquad 
\textbf{Zejian Xie}$^2$\qquad 
\textbf{Zhuoyang Song}$^2$\qquad  
\textbf{Songxin Zhang}$^2$\footnotemark[2]
\\
$^1$Zhejiang University \qquad
$^2$Southern University of Science and Technology\qquad \\
$^3$Beijing University of Technology \qquad 
$^4$Shanghai Artificial Intelligence Laboratory \qquad \\
$^5$Hong Kong Polytechnic University \qquad 
$^6$Qingdao University\\
{\tt\small 1471580435@qq.com\qquad xiaodayang@zju.edu.cn}
}

\begin{document}
\maketitle
\footnotetext[1]{*: equal controbution}
\footnotetext[2]{\dag: corresponding author}

\input{sec/0_abstract}    
\input{sec/1_intro}
\input{sec/2_relatedwork}

\input{sec/3_method}
\input{sec/4_experiment}

\input{sec/5_conclusion}

{
    \small
    \bibliographystyle{ieeenat_fullname}
    \bibliography{main}
}

\input{sec/X_suppl}

\end{document}

%% file: preamble.tex
\usepackage{color}
\usepackage{tabularray}
\usepackage{multirow}

\definecolor{cvprblue}{rgb}{0.21,0.49,0.74}
\usepackage[pagebackref,breaklinks,colorlinks,allcolors=cvprblue]{hyperref}

%% file: sec/0_abstract.tex
\begin{abstract}
 Vision-Language Models (VLMs) based on Mixture-of-Experts (MoE) architectures have emerged as a pivotal paradigm in multimodal understanding, offering a powerful framework for integrating visual and linguistic information. However, the increasing complexity and diversity of tasks present significant challenges in coordinating load balancing across heterogeneous visual experts, where optimizing one specialist's performance often compromises others' capabilities. To address task heterogeneity and expert load imbalance, we propose Astrea, a novel multi-expert collaborative VLM architecture based on progressive pre-alignment. Astrea introduces three key innovations: 1) A heterogeneous expert coordination mechanism that integrates four specialized models (detection, segmentation, classification, captioning) into a comprehensive expert matrix covering essential visual comprehension elements; 2) A dynamic knowledge fusion strategy featuring progressive pre-alignment to harmonize experts within the VLM latent space through contrastive learning, complemented by probabilistically activated stochastic residual connections to preserve knowledge continuity; 3) An enhanced optimization framework utilizing momentum contrastive learning for long-range dependency modeling and adaptive weight allocators for real-time expert contribution calibration. Extensive evaluations across 12 benchmark tasks spanning VQA, image captioning, and cross-modal retrieval demonstrate Astrea's superiority over state-of-the-art models, achieving an average performance gain of +4.7\%. This study provides the first empirical demonstration that progressive pre-alignment strategies enable VLMs to overcome task heterogeneity limitations, establishing new methodological foundations for developing general-purpose multimodal agents.
\end{abstract}

%% file: sec/1_intro.tex
\section{Introduction}
Visual-Language Models (VLMs), by integrating visual and linguistic modalities, have become a core paradigm for multimodal understanding tasks. VLMs aim to achieve complex understanding tasks through the simultaneous processing of image and text data. The key advantage of VLMs lies in their ability to unify multimodal and multitask scenarios within a single framework.

Existing VLM research mainly follows two paradigms: 1) constructing a unified multitask training framework \cite{chen2020uniter, su2019vl}, which achieves cross-task versatility through shared models and joint training, characterized by model simplicity and support for knowledge transfer between tasks; 2) utilizing task-specific expert models for feature fusion \cite{li2020oscar, zhang2021vinvl}, which optimizes task performance through specialized design and flexible integration, characterized by strong task-specificity and high modality complementarity.

However, both approaches face the challenge of balancing task heterogeneity and model generality. Specifically, while the former emphasizes model generality through a unified framework, it tends to encounter representation conflict issues in scenarios where cross-task correlations are insufficient. For instance, studies \cite{dou2024loramoe} have shown that when handling spatial localization and semantic description tasks simultaneously, due to significant differences in task objectives, the model can easily confuse its concentration on vision features, leading to performance degradation. This phenomenon is particularly evident in complex multitask settings, limiting the model's generalization ability. The latter approach designs different expert models for heterogeneous tasks, which alleviates task conflicts to some extent. However, due to the lack of generality among models, the efficiency of expert collaboration is low. Empirical studies \cite{zhou2025chatvlaunifiedmultimodalunderstanding, cai2024survey} indicate that multi-model architectures lead to parameter redundancy and knowledge-loss problems. 

To address these challenges, we propose a dynamic knowledge fusion Mixture of Expert (MoE) architecture. The core innovation lies in enhancing the collaborative capabilities between different experts through adaptive knowledge-sharing and task isolation mechanisms. 

Specifically, we design a coarse-to-fine pre-alignment strategy in the upstream training, and introduce a dynamic knowledge fusion module during downstream training to adaptively adjust the intensity of information exchange between experts, and protect task-specific knowledge through parameter isolation mechanisms.

In this paper, we propose \textbf{Astrea}, a multi-expert collaborative VLM architecture based on progressive pre-alignment, whose innovations are reflected in four dimensions: 1) \textbf{Heterogeneous Expert Collaboration Mechanism}: we select four fundamental task-specific expert models—Localization (spatial awareness), Segmentation (boundary parsing), Classification (semantic induction), and Captioning (context generation)—to construct an expert matrix that covers core elements of visual understanding; 2) \textbf{Progressive pre-alignment policy}: we align each expert with the VLM latent space, to reduce the training burden of the main model, and we activate information flow between experts via probabilistic sampling, which retains expert specificity while promoting knowledge transfer; 3) \textbf{Optimization Objective Enhancement}: we employ momentum contrastive learning to capture long-range dependencies, combined with a weight allocator to dynamically adjust the contribution of each expert. This approach aims to enhance both the generalization and collaboration efficiency of the model across diverse tasks.

In evaluations across 12 benchmark tasks, including VQA, image captioning, and cross-modal retrieval, Astrea significantly outperforms existing state-of-the-art (SOTA) models (with an average improvement of +4.7\%). This study is the first to demonstrate that through a progressive pre-alignment policy, VLMs can overcome the barriers of task heterogeneity, providing a new methodology for building universal multimodal agents.

Our main contributions are as follows:
\begin{itemize}
\item We propose \textbf{Astrea}, a MoE vision-language model, to address the challenge of balancing multi-task heterogeneity and model generalizability.
\item We introduce \textbf{a progressive pre-alignment policy} that designs a training scheme from coarse-grained to fine-grained, and by incorporating residual connections, effectively prevents knowledge forgetting, significantly enhancing model performance.
\item We propose a simple and efficient \textbf{dynamic feature fusion method} and further enhance model performance using \textbf{momentum contrastive learning}.
\item Our model achieves SOTA performance across multiple domains, and we will open-source our code.
\end{itemize}

%% file: sec/2_relatedwork.tex
\section{Related Work}

\subsection{Large Vision-Language Model}
Mainstream Vision-Language Models (VLMs) are typically based on the Transformer architecture of deep learning and achieve multimodal understanding by integrating visual and linguistic features. For example, the mPLUG \cite{li2022mplug}, the Qwen2-VL \cite{wang2024qwen2}, and the LLaVA-OneVision \cite{li2024llava} are all based on the Transformer architecture and incorporate pre-training along with self-supervised learning. mPLUG introduces an effective and efficient vision-language architecture with novel cross-modal skip connections, enabling time-consuming self-attention on the visual side through the creation of inter-layer shortcuts \cite{li2022mplug}. Qwen2-VL introduces the Naive Dynamic Resolution mechanism, enabling the model to dynamically process images of different resolutions into varying numbers of visual tokens. This allows the model to generate more efficient and accurate visual representations \cite{wang2024qwen2}. LLaVA-OneVision simultaneously pushes the performance boundaries of open-domain Large Multimodal Models (LMMs) across three key computer vision scenarios—single image, multiple images, and video—while enabling powerful transfer learning across different modalities and scenarios, resulting in the emergence of new capabilities \cite{li2024llava}. InternLM-XComposer-2.5 
 \cite{zhang2024internlm} and MoE-LLaVA \cite{lin2024moe} both demonstrate strong performance in enhancing the model's multitask adaptability and computational efficiency. InternLM-XComposer-2.5 supports long-context inputs and outputs, maintaining strong text-image understanding capabilities while also expanding the use of additional LoRA parameters for text-image synthesis \cite{zhang2024internlm}. MoE-LLaVA employs a mechanism of a mixture of experts, dynamically selecting the most relevant "expert" sub-models during inference based on the input. This significantly reduces computational costs while maintaining high performance \cite{lin2024moe}.

\subsection{Rethinking Image Representation}
Captioning, classification, detection, and segmentation tasks provide a comprehensive capture of visual information, as they encompass the core dimensions of image understanding-spatial and semantic aspects \cite{krizhevsky2012imagenet}. First, detection and segmentation tasks capture spatial information by providing the precise location of objects and pixel-level boundary separation. This enables the model to understand the arrangement, shape, and relative positioning of objects, which is crucial in scenes with multiple objects and complex backgrounds \cite{he2016deep, ren2016faster}. Secondly, classification and captioning tasks capture semantic information. Caption answers the questions that “what is happening” and “what is the picture as a whole”, offering a comprehensive description of the image through natural language and enabling the model to express itself across modalities \cite{xu2015show, anderson2018bottom}. Classification further addresses the question of “what are the objects in the picture” by identifying object categories, providing more semantic understanding of the image \cite{deng2009imagenet}. Moreover, these tasks complement each other to form a complete visual understanding framework: detection and segmentation help the model build spatial awareness, while classification and captioning enable the model to grasp semantic associations \cite{redmon2016you, lin2014microsoft}. Therefore, these tasks collectively create a effective system for capturing visual information. By covering aspects like spatial location, shape boundaries, category recognition, and semantic description, they provide the model with multi-level information, enabling it to understand image content more accurately and comprehensively.

%% file: sec/3_method.tex
\graphicspath{{Figure/}}
\begin{figure*}[t]
  \centering
  \includegraphics[width=\linewidth]{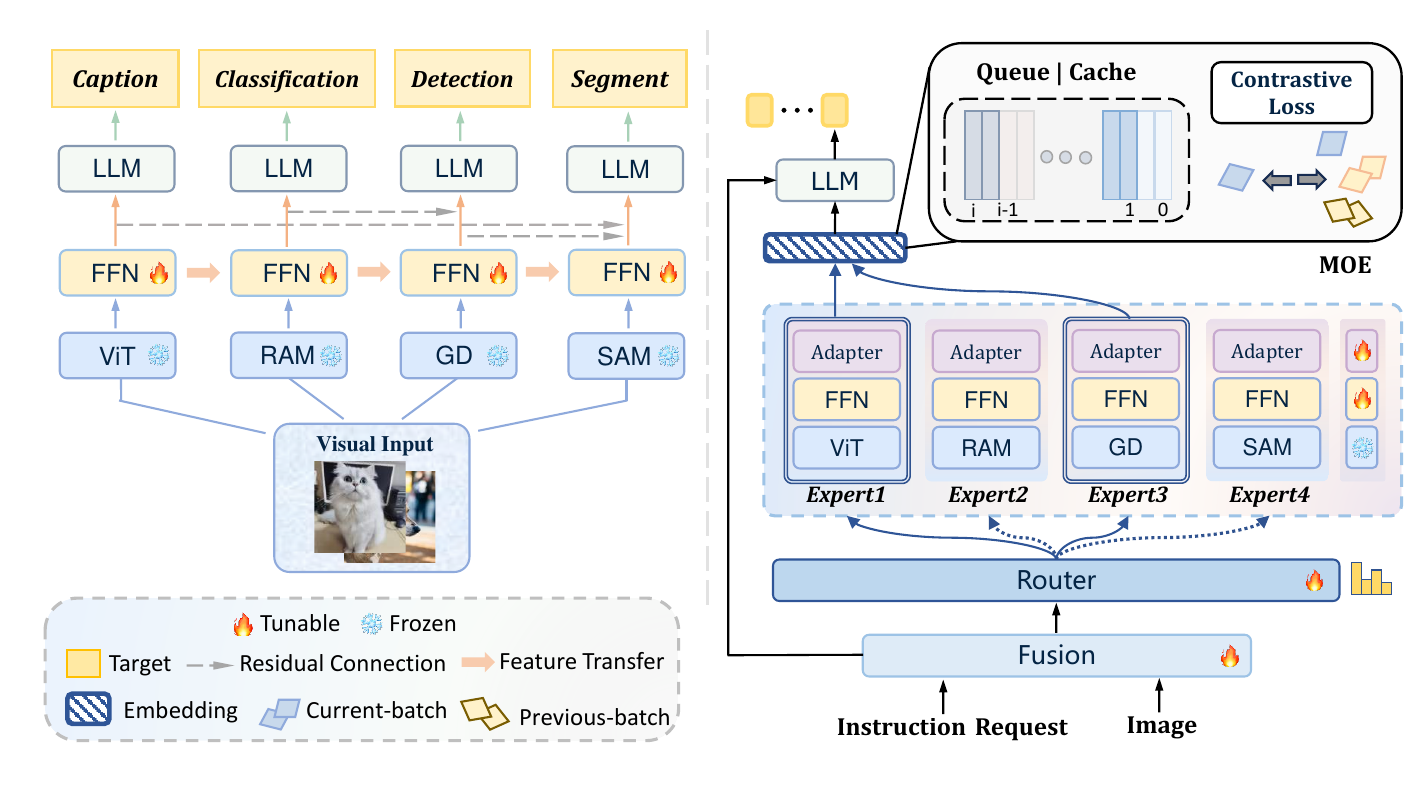}
  \vspace{-5pt}
  \caption{The pipeline of the Astrea. The left panel illustrates the four stages of pre-alignment, with the Instruct phase not explicitly shown. Note that the target labels simply highlight the emphasized points; in reality, the target scope is gradually expanded. The right panel presents the main model, which is initialized with the outcomes from the left panel.}
  \label{fig:1}
\end{figure*}

\section{Method}
Our task is to obtain a unified MoE-based model that can solve the problem of heterogeneous expert collaboration, through two-stage training. As shown in Fig. \ref{fig:1}, our training is conducted in two stages. The first is the alignment stage, where the model progresses from coarse-grained to fine-grained understanding, as detailed in Sec. \ref{Sec:3.1.1}. During this process, we employ residual connections to prevent forgetting, which will be detailed in Sec. \ref{Sec:3.1.2}. Since the VLM has already been adapted to the tasks during the alignment stage, its training load is significantly reduced in the main model training phase. Instruct Tuning can be detailed in Sec. \ref{Sec:3.1.3}. We use momentum learning to enhance feature accuracy, as discussed in detail in Sec. \ref{Sec:3.2.1}. Most importantly, we propose a novel MOE network that uses dynamic weight allocation, as elaborated in Sec. \ref{Sec:3.2.2}.

\subsection{Alignment}
In this stage, our goal is to align the large language model with various models, completing the initialization for the Astrea module.
\subsubsection{Progressive Training}\label{Sec:3.1.1}
From the perspective of visual information representation, the semantics, spatial positions, and relationships between objects can effectively capture the essential content of an image. Therefore, in designing the feature extraction strategy, we primarily incorporate these three types of information. Our training process follows a progressive approach from coarse-grained to fine-grained, beginning with a global perspective and gradually moving to more detailed, local information. For the interpretation of granularity, further discussions are presented in the supplementary. This stepwise refinement process helps the model establish a global understanding of the overall visual information in the early stages, then progressively capture more detailed features. During model training, we use a feedforward neural network (FFN) \cite{lecun2015deep} to process features, aiming to align features from different sources to the same dimension, which also enhances the model’s flexibility.

Given an RGB image $ I\in \mathbb{R}^{H \times W \times 3} $, where $ H $ and $ W $ are the origin resolution. The model processes input images to obtain a visual token sequence $ Z = [z_1, z_2, \dots, z_P] \in \mathbb{R}^{P \times D} $, where $ P $ represents the sequence length of visual tokens, $ D $ represents the hidden size of VLM. Similarly, the text undergoes a word embedding layer and is projected to obtain the sequence tokens $ T = [t_1, t_2, \dots, t_N] \in \mathbb{R}^{N \times D} $, where $ N $ represents the sequence length of text tokens.

We optimize the output of VLM through a generative loss in an auto-regressive manner. Given an image and text, MoE-LLaVA \cite{lin2024moe} generates the output sequence $Y = [y_1, y_2, \dots, y_K] \in \mathbb{R}^{K \times D}$ by progressively generating each element, where $K=P+N$ represents the length of the output sequence. The formula is:
\begin{equation}
    \mathcal{L}_{\text{reg}} = - \sum_{i=1}^{N} \log p_{\theta} \left( Y^{[P+i]} \mid Z, T^{[i-1]} \right)
\end{equation}
where $\theta$ is a trainable parameter and we only calculate the loss for the generated text.

\subsubsection{Random Residual Connection}\label{Sec:3.1.2}
During the pre-alignment phase, to prevent the model from forgetting knowledge learned from previous tasks when training on new targets, we introduce the concept of residual connections. Specifically, we randomly introduce residual connections between tasks, enabling the model to flexibly leverage knowledge from previous tasks and reduce forgetting. This residual mechanism \cite{he2016deep} introduces a degree of coherence between tasks, thereby strengthening the model’s ability to adapt to multiple tasks. Inspired by memory bank \cite{he2020momentum, noroozi2016unsupervised}, we create a cache $C \in \mathbb{R}^{n \times m}$ for each model, which is used to record the features of each image, where \textit{n} is the number of models and \textit{m} is the number of images, then the embedding of the \textit{j}-th image in the \textit{i}-th cache can be expressed as $C_{i}^{j}$. This approach not only avoids recomputing feature representations from previous tasks but also enables efficient information sharing across multiple tasks. When proceeding to the \textit{k}-th model, the feature of image \textit{j} processed by the feed-forward network (FFN) of model \textit{k} is notated as $f_{kj}$, the feature input to the large language model (VLM) is then expressed as:
\begin{equation}
    f_{kj} ^\prime= \text{Softmax}\left(\frac{Q K^T}{\sqrt{d}}\right) \cdot V
\end{equation}
\begin{equation}
    Q=\mathcal{F}_q (\gamma \cdot \sum_{i=1}^{k-1}  C_{i}^{j} \cdot p_i), 
\end{equation}
\begin{equation}
    K=\mathcal{F}_k(f_{kj}) , V=\mathcal{F}_v(f_{kj})
\end{equation}
where $p_i \in \{0,1\}$ is a binary variable used to control whether a residual connection is added with a certain probability, and $\gamma$ is a weight parameter for the residual connection. The functions $\mathcal{F}_q$, $\mathcal{F}_k$, and $\mathcal{F}_v$ are mapping functions that convert features into representations of queries, keys, and values, respectively. After training the \textit{k}-th model, we freeze its parameters. At the same time, we will update the feature cache $C_k$ of the current model if multi-round pretraining is required. This progressive freezing and caching strategy allows the model to continually learn new tasks without losing knowledge from previous ones.

\subsubsection{Instruct Tuning}\label{Sec:3.1.3}
To begin with, the dataset needs to be constructed. The alignment process labels for each stage contain a summary of the labels from previous stages. Taking RAM \cite{mnih2014recurrent} as an example, the instruction is set to “caption, recognization”. Other visual experts also undergo similar processing.

As shown in the left panel of Fig. \ref{fig:1}, FFN and residual connection transmit the features extracted from the completed training task to the current training task. The training proceeds sequentially from left to right, and each time a new task is introduced, the model considers not only the target of the current task but also that of the previous task. This stage-wise learning mechanism, combined with the consideration of prior task tokens, helps to prevent knowledge forgetting and ensures that the model retains the learning outcomes of previous tasks while undertaking new ones.

\subsection{Architecture of Astrea}\label{Sec:3.2.1}
At this stage, the VLM is adapted with multimodal understanding capabilities.
\subsubsection{Momentum Contrast Learning}
We choose contrastive learning \cite{chen2020simple} as our learning strategy, which forces the model to capture the essential semantic differences between samples, rather than trivial surface features. Empirically, increasing the number of negative samples significantly enhances model training, as more negative samples help the model better distinguish between positive and negative samples. However, Contrastive learning performs poorly when the batch size is small. Besides, due to computational resource constraints, it is not feasible to simply increase the batch size to add more negative samples. To address this issue, we employ momentum contrast learning \cite{he2020momentum}. Specifically, we maintain a queue to store sample representations from past batches, effectively increasing the number of negative samples without needing to enlarge the current batch size. The current batch is added to the queue, while the oldest batch in the queue is removed. This approach allows us to flexibly set the queue size without being constrained by batch size.

Let the batch size be denoted as $B$, and the queue as $V \in \mathbb{R}^{N \times D}$, where $N$ is the queue size and $D$ is the feature dimension. First, the embeddings of the current batch are updated into the queue, while the other embeddings in the queue are updated by the momentum encoder. Thus, the momentum contrastive learning loss is calculated as follows:
\begin{equation}
    S = \frac{V \cdot V^T}{\tau} \in \mathbb{R}^{N \times N}
\end{equation}
\begin{equation}
    r_i=\frac{e^{S_{ii}}}{\sum_j e^{S_{ij}}}; c_i=\frac{e^{S_{ii}}}{\sum_j e^{S_{ji}}}
\end{equation}
\begin{equation}\label{eq:7}
    \mathcal{L}_{\text{moco}} = -\frac{1}{N} \sum_{i=1}^{B} \left[ \lambda \cdot \log\left(r_i\right) + (1 - \lambda) \cdot \log\left(c_i\right) \right]
\end{equation}
where $\tau$ is a scaling factor to prevent overfitting, while $\lambda$ is a temperature hyperparameter that controls the contributions of contrastive learning from the two perspectives of rows and columns.

Let the current $\theta_{E}$ and $\theta_{M}$ denote the parameters of the current encoder and the momentum encoder, respectively. Before the end of the current batch, update the momentum encoder according to Eq. \ref{eq:7}:
\begin{equation}\label{eq:8}
    \theta_{M}=m\cdot \theta_{M}+(1-m)\cdot \theta_{E}
\end{equation}
where $m \in [0,1)$ is a momentum coefficient. Only the parameters $\theta_{E}$ are updated by back-propagation. The momentum update in Eq. \ref{eq:8} makes $\theta_{M}$ evolve more smoothly than $\theta_{E}$. In this way, though the keys in the queue are encoded by different encoders across different batches, the variation among these encoders can be minimized. Our design considers both the relative relationships among samples within the same batch and those between samples across different batches. This approach effectively increases the number of negative samples, thereby enhancing training effectiveness and improving the model’s generalization ability.

\subsubsection{Better Trade-offs}\label{Sec:3.2.2}
Traditional models tend to forget knowledge learned from previous tasks when training on new tasks, which limits their performance in multi-task learning. To address this issue, we introduce the concept of MoE \cite{fedus2022switch}, using independent expert modules to retain the capabilities required for each task. This design allows each model to focus on a specific task without losing knowledge from previous tasks when learning new ones, effectively preserving knowledge across tasks. Since the adaptation of the VLM is already completed in the alignment stage, the training load on the VLM is significantly reduced in the main model stage.

The router is a linear layer that predicts the probability of the tokens being assigned to each expert, which can be denoted as $\omega_i$. Since we ultimately only mix the decisions of the top $k$ experts, we start by sorting the weights $w_i$ in descending order. We then set the smallest $n-k$ weights to zero and apply softmax to polarize the distribution, ensuring that each task tends to be handled by a specific expert as much as possible. We formulate as:
\begin{equation}
    \mathcal{P}_i = \frac{e^{\omega_i}}{\sum_j^n e^{\omega_j}}
\end{equation}

In addition, we introduce an innovative approach by using an adapter for weight redistribution, reducing the likelihood of decision errors. With each input, the adapter dynamically allocates weights to different expert modules based on the characteristics of the data, thereby determining the contribution of each expert to the final output. Let the output feature of the FFN be denoted as $f_i$, then the output of the MoE can be expressed as:
\begin{equation}\label{eq:10}
    f=\sum_{i=1}^{k} \mathcal{P}_{i} \cdot \mathcal{F}_i(f_i)
\end{equation}
where $\mathcal{F}_i$ is the adapter mentioned in the Fig. \ref{fig:1}, which is realized through an MLP network. The parameter $k$ denotes the number of experts involved in the decision-making process. This setup implies that not all experts participate in each decision; rather, the model adaptively selects the $k$ (default 3) most likely experts to contribute. 

Ultimately, the total loss is expressed as:
\begin{equation}
    \mathcal{L}_{total}=\mathcal{L}_{\text{reg}}+\mu \cdot \mathcal{L}_{\text{moco}}
\end{equation}
where $\mathcal{L}_{reg}$ is computed in the same way as in the alignment phase. $\mu$ is a hyperparameter that regulates the contribution of contrastive learning.

%% file: sec/4_experiment.tex
\begin{table*}[t] 
\centering
\renewcommand{\arraystretch}{1.5}
\resizebox{\textwidth}{!}{ 
\begin{tblr}{
  row{1} = {c},
  row{3} = {c},
  row{9} = {c},
  cell{1}{1} = {c=2}{},
  cell{1}{3} = {c=4}{},
  cell{1}{7} = {c=4}{},
  cell{1}{11} = {c=4}{},
  cell{1}{15} = {c=4}{},
  cell{3}{1} = {c=18}{},
  cell{9}{1} = {c=18}{},
  vline{2,4,8,12} = {1}{},
  vline{3,7,11,15} = {2,4-8,10-13}{},
  hline{1-4,9-10,14} = {-}{},
  row{8,13} = {bg={yellow!20}}
}
Model                    &      & General         &               &               &               & Knowledge     &               &               &               & OCRChart      &               &               &               & Vision-Centric &               &               &               \\
Method                   & \rotatebox{75}{\scalebox{0.7}{$\bm{VIs Tok.}$}} & \rotatebox{75}{\scalebox{0.6}{$\bm{MME^P}$}} & \rotatebox{75}{\scalebox{0.7}{$\bm{MMB}$}}  & \rotatebox{75}{\scalebox{0.6}{$\bm{SEED^I}$}} & \rotatebox{75}{\scalebox{0.7}{$\bm{GQA}$}}  & \rotatebox{75}{\scalebox{0.7}{$\bm{SQA^I}$}}   & \rotatebox{75}{\scalebox{0.7}{$\bm{MMMU^V}$}} & \rotatebox{75}{\scalebox{0.7}{$\bm{MathVista^M}$}} & \rotatebox{75}{\scalebox{0.7}{$\bm{AI2D}$}} & \rotatebox{75}{\scalebox{0.7}{$\bm{ChartQA}$}}  & \rotatebox{75}{\scalebox{0.7}{$\bm{OCRBench}$}} & \rotatebox{75}{\scalebox{0.7}{$\bm{TextVQA}$}} & \rotatebox{75}{\scalebox{0.7}{$\bm{DocVQA}$}} & \rotatebox{75}{\scalebox{0.7}{$\bm{MMVP}$}}           & \rotatebox{75}{\scalebox{0.7}{$\bm{RealworldQA}$}} & \rotatebox{75}{\scalebox{0.7}{$\bm{{CV\text{-}Bench}^{2D}}$}} & \rotatebox{75}{\scalebox{0.7}{$\bm{CV\text{-}Bench^{3D}}$}} \\
Base LLM: Vicuna-1.5-13B &      &                 &               &               &               &               &               &               &               &               &               &               &               &                &               &               &               \\
Mini-Gemini-HD-13B       & 2880 & 1597.0          & 68.6          & 70.6          & 63.7          & 71.9          & 37.3          & 37.0          & 70.1          & 56.6          & 46.6          & 70.2          & 69.8          & 19.3           & 57.5          & 53.6          & 67.3          \\
LLaVA-NeXT-13B           & 2880 & 1575.0          & 70.0          & 65.6          & 65.4          & 73.5          & 36.2          & 35.1          & 70.0          & 62.2          & 51.4          & 67.1          & 70.9          & 36.0           & 59.1          & 62.7          & 65.7          \\
LLaVA-1.5-13B            & 576  & 1531.3          & -             & -             & 63.3          & -             & -             & -             & -             & -             & -             & 61.3          & -             & -              & -             & -             & -             \\
Cambrian-1-13B           & 576  & 1610.4          & 75.7          & 74.4          & 64.3          & \textbf{79.3} & 40.0          & 48.0          & 73.6          & 73.8          & 61.9          & 72.8          & \textbf{76.8} & 41.3           & 63.0          & 72.5          & 71.8          \\
Astrea-13B               & 576  & \textbf{1655.2} & \textbf{77.2} & \textbf{76.1} & \textbf{65.9} & 78.2          & \textbf{42.5} & \textbf{49.3} & \textbf{74.6} & \textbf{75.5} & \textbf{63.4} & \textbf{73.8} & 75.5          & \textbf{44.6}  & \textbf{65.0} & \textbf{74.1} & \textbf{73.0} \\
Base LLM: Hermes2-Yi-34B &      &                 &               &               &               &               &               &               &               &               &               &               &               &                &               &               &               \\
Mini-Gemini-HD-34B       & 2880 & 1659.0          & 80.6          & 75.3          & 65.8          & 77.7          & 48.0          & 43.4          & 80.5          & 67.6          & 51.8          & 74.1          & 78.9          & 37.3           & 67.           & 71.5          & 79.2          \\
LLaVA-NeXT-34B           & 2880 & 1633.2          & 79.3          & 75.9          & \textbf{67.1} & 81.8          & 46.7          & 46.5          & 74.9          & 68.7          & 54.5          & 69.5          & 78.1          & 47.3           & 61.0          & 73.0          & 74.8          \\
Cambrian-1-34B           & 576  & 1689.3          & 81.4          & 75.3          & 65.8          & \textbf{85.6} & 49.7          & 53.2          & 79.7          & 75.6          & 60.0          & 76.7          & 75.5          & \textbf{52.7}  & 67.8          & 74.0          & 79.7          \\
Astrea-34B               & 576  & \textbf{1749.5} & \textbf{83.6} & \textbf{76.9} & 66.7          & 88.2          & \textbf{51.5} & \textbf{55.4} & \textbf{82.4} & \textbf{78.2} & \textbf{63.0} & \textbf{79.2} & \textbf{80.6} & 51.4           & \textbf{69.8} & \textbf{75.5} & \textbf{82.2} 
\end{tblr}
}
\caption{Comparison of our Astrea with other leading MLLM frameworks. Using only 576 visual tokens, Astrea has demonstrated competitive performance in multiple benchmark tests.}
\label{table1}
\end{table*}
\section{Experiment}

\begin{table*}
\centering
\resizebox{\textwidth}{!}{%
\begin{tabular}{l|cccc|cc|cl} 
\hline
\multicolumn{1}{c|}{\multirow{3}{*}{Model}}    & \multicolumn{4}{c|}{MC-VQA}                                          & \multicolumn{2}{c|}{VC}           & \multicolumn{2}{c}{OE-VQA}                           \\ 
\cline{2-9}
\multicolumn{1}{c|}{}                          & EgoSchema     & Perception-Test & MVBench       & VideoMME           & \multicolumn{2}{c|}{MSVC (Score)} & MSVD              & \multicolumn{1}{c}{ActivityNet}  \\
\multicolumn{1}{c|}{}                          & (Acc)         & (Acc)           & (Acc)         & (Acc)              & correctness   & detailedness      & (Acc./Score)      & (Acc./Score)                     \\ 
\hline
\multicolumn{9}{c}{Base LLM}                                                                                                                                                                                     \\ 
\hline
Video-LLaVA (7B)                               & 38.4          & 44.3            & 41.0          & 39.9/41.6          & 1.85          & 2.05              & 70.7/3.9          & 45.3/3.3                         \\
LLaVA-NeXT-Video (7B)                          & 43.9          & 48.8            & 46.5          & -                  & 2.40          & 2.52              & 67.8/3.5          & 53.5/3.2                         \\
VideoLLaMA2 (7B)                               & 51.7          & 51.4            & 54.6          & 47.9/50.3          & 2.53          & 2.59              & 70.9/3.8          & 50.2/3.3                         \\
VideoLLaMA2.1 (7B)                             & 53.1          & \textbf{54.9}   & 57.3          & 54.9/56.4          & \textbf{2.87} & 2.81              & 70.6/3.8          & 53.0/3.4                         \\
\rowcolor[rgb]{0.851,0.953,0.992} Astrea (13B) & \textbf{57.3} & 53.6            & \textbf{57.7} & \textbf{56.2/57.7} & 2.69          & \textbf{2.82}     & \textbf{72.0/3.9} & \textbf{53.1/3.4}                \\ 
\hline
\multicolumn{9}{c}{Large LLM}                                                                                                                                                                                    \\ 
\hline
LLaVA-NeXT-Video (32B)                         & 60.9          & -               & -             & 60.2/63.0          & -             & -                 & -                 & 54.3/-                           \\
PLLaVA (34B)                                   & -             & -               & 58.1          & -                  & -             & -                 & -                 & 60.9/-                           \\
VideoLLaMA2 (72B)                              & 63.9          & 57.9            & 62.0          & 61.4/63.1          & 2.71          & 2.67              & 71.0/3.8          & 55.2/3.4                         \\
\rowcolor[rgb]{0.851,0.953,0.992} Astrea (34B) & \textbf{66.5} & \textbf{60.5}   & \textbf{64.2} & \textbf{62.5/64.8} & \textbf{2.96} & \textbf{3.13}     & \textbf{73.3/4.0} & \textbf{58.2/3.6}                \\
\hline
\end{tabular}}
\caption{The main results of Multiple-choice Video Question Answering (MC-VQA), Video Captioning (VC), and Open-ended Video Question Answering (OE-VQA) benchmarks. Astrea demonstrates advantages in the field of multi-modal large language models, excelling in accuracy across multiple benchmarks. It is competitive, and often superior, to models of various parameter scales. With strong adaptability, Astrea performs well on both simple and complex tasks, making it a standout in the multi-modal large language model landscape.}
\label{table:2}
\end{table*}

\subsection{Implement Details}
We follow the LLAVA-Next \cite{liu2024llava} framework for multi-resolution settings, which divides high-resolution images into multiple low-resolution patches. The AdamW optimizer is employed, with parameters $\beta_1 = 0.9$ and $\beta_2 = 0.98$, and a weight decay rate of 0.05. A cosine learning rate scheduler is used during training, with a peak learning rate of 1e-4 and a linear warm-up rate of 15\%.

For the text encoder, we evaluate both RAM and RAM++. For object detection, we test Grounding-DINO and Grounding-DINO-1.5, while for segmentation, we explore SAM and SAM2. In the global information extraction module, various models are assessed, including OpenAI CLIP ViT-L/14@336, EVA-CLIP-02 ViT-L/14@336, SigLIP ViT-SO400M/14@384, and DINOv2 ViT-L/14@336. After pre-alignment, we select the most effective combination: SigLIP, RAM++, Grounding-DINO-1.5, and SAM2.
\vspace{-0.1cm}
The training is conducted on 48$\times$8 A800 GPUs, organized into four pre-alignment stages with step counts of 2000k, 500k, 200k, and 200k, respectively. For instruction fine-tuning, the number of training steps is set to 500k, with a uniform batch size of 128.

Details about the dataset and instructions will be shown in Supplementary Material.

\graphicspath{{Figure/}}
\begin{figure*}[t]
  \centering
  \includegraphics[width=\linewidth]{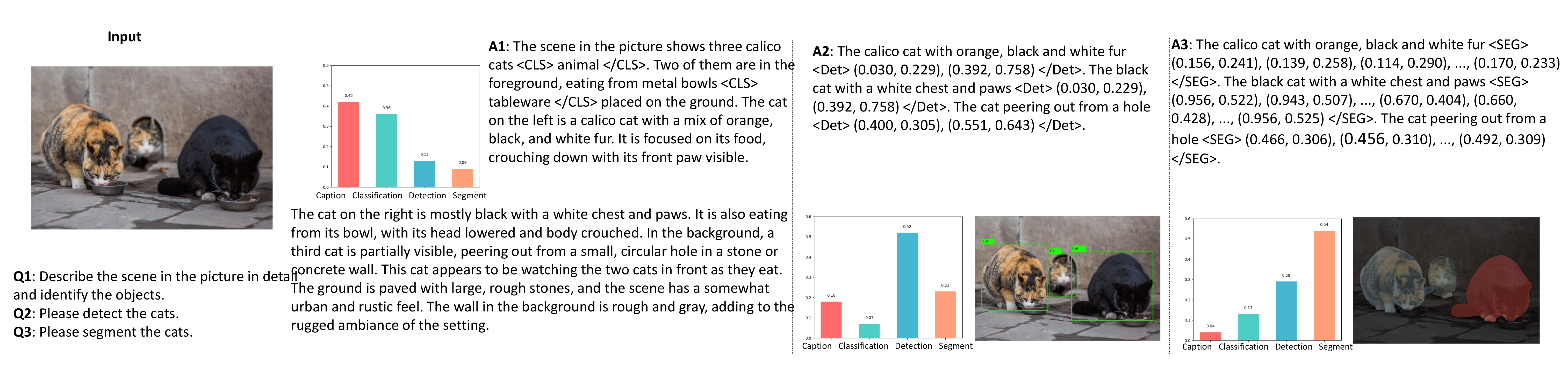}
  \caption{The demo of the QA task. The bar chart shows the contribution probabilities of the four models output by the router.}
  \label{fig:2}
\end{figure*}

\subsection{Metrics}
For images, we use a suite of commonly used benchmarks \cite{chang2024survey, ge2023planting, hiippala2021ai2d, hudson2019gqa, liu2025mmbench, liu2023hidden, lu2022learn, lu2023mathvista, masry2022chartqa, singh2019towards, tong2024eyes, yue2024mmmu}. In detail, for VQA tasks, performance is evaluated by comparing the model's responses with the ground-truth answers, and the Top-1 accuracy is reported. For localization ability, a prediction is considered correct if the Intersection over Union (IOU) between the predicted bounding box and the ground truth exceeds 0.5, and the accuracy is reported accordingly. In addition, we conducted a comprehensive evaluation across multiple toolkits to thoroughly assess the multimodal perception and conversational capabilities. The toolkits cover open-ended responses and factual accuracy assessments. 

For videos, we use GPT-3.5 \cite{fang2023data} to evaluate the accuracy (whether the answer is true/false) and quality (score ranging from 0 to 5) of the model's responses. Additionally, we adopted the video generation performance benchmark (VCG Score) introduced by VideoChatGPT \cite{dosovitskiy2020image}, which typically involves longer responses. This benchmark covers five aspects of video understanding: correctness of information, detail orientation, context understanding, temporal understanding, and consistency. The generation quality is also evaluated using the GPT-3.5 model.

\subsection{Main Result}
From the Fig. \ref{fig:2}, we can observe that when the task type is determined, the correct expert model corresponding to it contributes the most, e.g., the caption task corresponds to ViT. This alignment with logical expectations confirms that our MOE strategy is effective. From the Tab. \ref{table1} and Tab. \ref{table:2}, it can be observed that Astrea significantly outperforms other models across multiple tasks, particularly in the $Knowledge$ and $OCR \& Chart$ categories, which generally demand higher levels of cross-modal understanding and collaboration. The pre-alignment process enables Astrea to ensure coordination among different modules of the model before formal training, thereby reducing the burden during main model training. Additionally, as pre-alignment steps increase, the model can quickly adapt to simple $VQA$ tasks during the instruction fine-tuning phase. However, for complex visual reasoning dialogue tasks, effective adaptation requires thorough visual-language alignment beforehand.

In image comprehension tasks, the Astrea model demonstrates a significant advantage. On the General category datasets, including $MME^P$, $MMB$, $SEED$, and $GQA$, Astrea outperforms other models, especially excelling on the $MMB$ and $SEED$ datasets, indicating its multi-expert and dynamic weight allocation strategy’s strong accuracy in handling general knowledge and everyday information tasks.
In the Knowledge category, across datasets like $SQA^I$, $MMMU$, $MathVista$, and $AI2D$, Astrea also achieves outstanding performance, highlighting its impressive understanding and application capabilities in specialized fields like science and mathematics. This strength is attributed to its multi-expert structure and dynamic task selection capability.
In the $OCR \& Chart$ category, on datasets such as $ChartQA$, $OCRBench$, and $DocVQA$, Astrea consistently scores high, demonstrating the visual module’s efficiency in text recognition and chart information comprehension.
Similarly, in the Vision-Centric category, Astrea excels on datasets like $RealworldQA$, $CV\text{-}Bench2D$, and $CV\text{-}Bench3D$, scoring especially well on $CV\text{-}Bench3D$, showcasing its robust capabilities in spatial and depth information understanding.
Overall, these results suggest that Astrea exhibits outstanding performance across general, knowledge-based, OCR, and vision-centric tasks.

Astrea also demonstrates excellent performance in video understanding tasks. In the multiple-choice video QA ($MC\text{-}VQA$) task, Astrea-34B achieves outstanding accuracy on datasets such as $EgoSchema$, $Perception\text{-}Test$, $MVBench$, and $VideoMME$, indicating its strong comprehension and reasoning abilities in multiple-choice video question answering.
In the video captioning ($VC$) task, Astrea performs exceptionally well on the $MSVC$ dataset, surpassing other models ,particularly on the correctness and detailedness metrics, which highlights its advantage in generating precise and detailed descriptions.
Furthermore, in the open-ended video question answering ($OE\text{-}VQA$) task, Astrea-34B also leads in performance on the $MSVD$ and $ActivityNet$ datasets. These results demonstrate Astrea’s comprehensive and superior capabilities in video understanding tasks, showing its ability to accurately answer questions and generate content-rich descriptions.

\renewcommand{\arraystretch}{0.85} 
\begin{table}[t]
\resizebox{0.48\textwidth}{!}{
\centering
\renewcommand{\arraystretch}{0.4} 
\begin{tblr}{
   hline{1-2,7} = {-}{},
   hline{2} = {0.5pt},
   hline{1,7} = {0.55pt},
   rowsep=0.8pt
}
\rotatebox{53}{\scalebox{0.67}{$\bm{K}$}} & \rotatebox{53}{\scalebox{0.67}{$\bm{MME^P}$}}  & \rotatebox{53}{\scalebox{0.67}{$\bm{SQA^I}$}} & \rotatebox{53}{\scalebox{0.67}{$\bm{ChartQA}$}} & \rotatebox{53}{\scalebox{0.67}{$\bm{MMVP}$}} & \rotatebox{53}{\scalebox{0.67}{$\bm{MVBench}$}} & \rotatebox{53}{\scalebox{0.67}{$\bm{MSVD}$}} \\
Top-1        & 1668.5 & 86.0    & 76.5    & 47.5 & 61.7    & 71.5 \\
Top-2        & 1733.6 & 87.8  & \textbf{78.4}    & \textbf{51.8 }& 63.9    & 72.7 \\
Top-3        & \textbf{1749.5} & \textbf{88.2}  & 78.2    & 51.4 & \textbf{64.2}    & 73.3 \\
Top-4        & 1713.6 & 87.4  & 77.6    & 51.1 & 63.6    & \textbf{73.6} \\
Average          & 1702.2 & 86.6  & 77.0      & 50.3 & 62.3    & 71.8 
\end{tblr}
}
\caption{Ablation study on the number of experts involved in decision-making. k represents the top-k models with the highest probabilities that participate in the decision, as defined in Eq. \ref{eq:10}.}
\label{table:3}
\end{table}

\begin{table}[t]
\resizebox{0.48\textwidth}{!}{
\centering
\begin{tblr}{
   hline{1-2,5} = {-}{},
   hline{2} = {0.5pt},
   hline{1,5} = {0.55pt},
}
\rotatebox{65}{\scalebox{0.67}{$\bm{Arch.}$}} & \rotatebox{65}{\scalebox{0.67}{$\bm{MME^P}$}}  & \rotatebox{65}{\scalebox{0.67}{$\bm{SQA^I}$}} & \rotatebox{65}{\scalebox{0.67}{$\bm{ChartQA}$}} & \rotatebox{65}{\scalebox{0.67}{$\bm{MMVP}$}} & \rotatebox{65}{\scalebox{0.67}{$\bm{MVBench}$}} & \rotatebox{65}{\scalebox{0.67}{$\bm{MSVD}$}} \\
w/o Residual        & 1701.5 & 87.6    & 77.5    & 50.1 & 63.3    & 72.4 \\
w/o Contrast        & 1683.3 & 87.4  & 76.8    & 50.5 & 63.6    & 72.1 \\
Astrea        & \textbf{1749.5} & \textbf{88.2}  & \textbf{78.2}    & \textbf{51.4} & \textbf{64.2}    & \textbf{73.3} \\
\end{tblr}
}
\caption{Ablation study on the residual and contrastive learning.}
\label{table:4}
\end{table}

\graphicspath{{Figure/}}
\begin{figure*}[t]
  \centering
  \includegraphics[width=\linewidth]{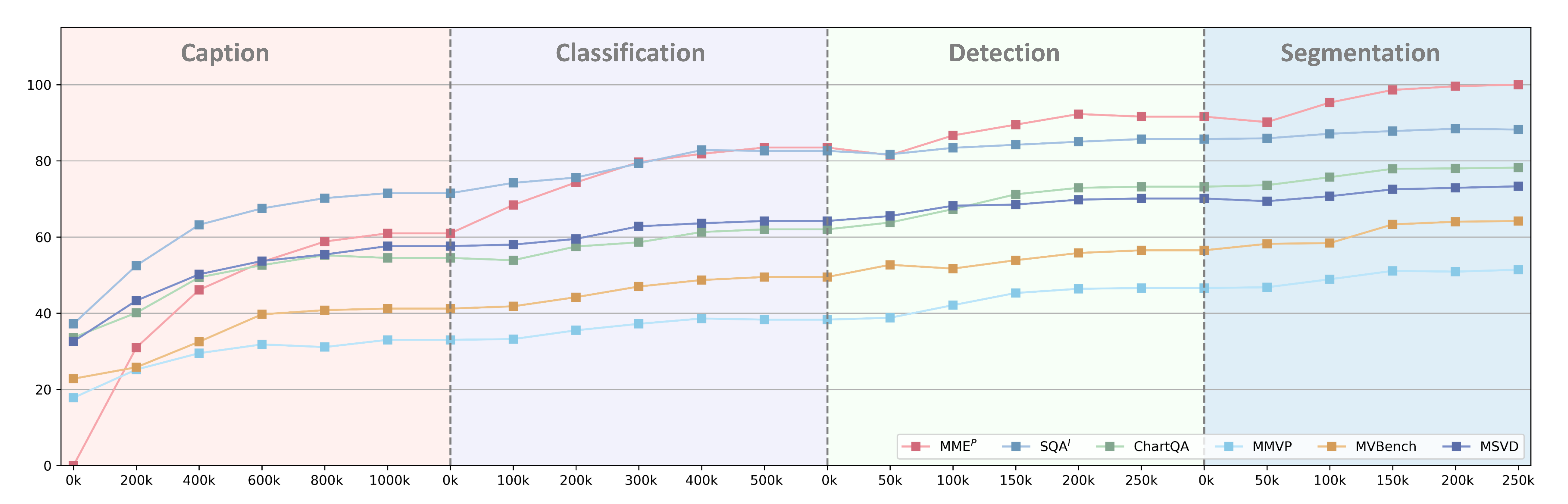}
  \caption{The trend of various metrics for different subjects across training steps during the alignment stage. The $MME^P$ results are normalized to a scale of 0-100. The alignment proceeds sequentially through  the Caption, Classification, Detection, and Segmentation stages.}
  \label{fig:3}
\end{figure*}

\begin{table}[t]
\definecolor{mycolor1}{rgb}{0.910, 0.973, 0.961} 
\definecolor{mycolor4}{rgb}{0.922, 0.961, 0.984} 
\definecolor{mycolor3}{rgb}{0.996, 0.976, 0.906}
\definecolor{mycolor2}{rgb}{0.957, 0.925, 0.969}
\centering
\resizebox{0.48\textwidth}{!}{ 
\begin{tblr}{
  hline{1-2,8} = {-}{},
  hline{2} = {0.45pt},
  hline{1,8} = {0.55pt},
  row{2-4} = {bg=mycolor1},       
  row{5} = {bg=mycolor2},
  row{6} = {bg=mycolor3},
  row{7} = {bg=mycolor4}
}
\rotatebox{65}{\scalebox{0.73}{$\bm{Encoder}$}} & \rotatebox{65}{\scalebox{0.71}{$\bm{MME^P}$}}  & \rotatebox{65}{\scalebox{0.71}{$\bm{SQA^I}$}} & \rotatebox{65}{\scalebox{0.71}{$\bm{ChartQA}$}} & \rotatebox{65}{\scalebox{0.71}{$\bm{MMVP}$}} & \rotatebox{65}{\scalebox{0.71}{$\bm{MVBench}$}} & \rotatebox{65}{\scalebox{0.71}{$\bm{MSVD}$}} \\
OpenAI CLIP  & 1683.3 & 87.1  & 76.9    & 49.2 & 62.8    & 71.9 \\
EVA-CLIP-02  & 1725.5 & 87.9  & 76.7    & 49.6 & 63.3    & 72.4 \\
DINOv2       & 1703.6 & 87.4  & 78.5    & 51.2 & 63.9    & 72.6 \\
RAM                       & 1722.5 & 88.0  & 77.7    & 50.8 & 63.7    & 72.8 \\
Grounding-DINO          & 1717.2 & 87.8  & 77.5    & 51.1 & 63.5    & 73.0   \\
SAM                      & 1695.5 & 88.1  & 77.2    & 50.9 & 64.0    & 72.7 
\end{tblr}
} 
\caption{Ablation study in the alignment stage by replacing each stage’s extractor with alternative models. The tests are conducted with a 34B LLM. The top three are alternative ViT models, and the optimal model selection results are shown in Tab. \ref{table1}.}
\label{table:5}
\end{table}

\begin{table}[t]
\resizebox{0.48\textwidth}{!}{
\centering
\begin{tblr}{
   hline{1-2,5} = {-}{},
   hline{2} = {0.5pt},
   hline{1,5} = {0.55pt},
}
\rotatebox{65}{\scalebox{0.67}{$\bm{Order}$}} & \rotatebox{65}{\scalebox{0.67}{$\bm{MME^P}$}}  & \rotatebox{65}{\scalebox{0.67}{$\bm{SQA^I}$}} & \rotatebox{65}{\scalebox{0.67}{$\bm{ChartQA}$}} & \rotatebox{65}{\scalebox{0.67}{$\bm{MMVP}$}} & \rotatebox{65}{\scalebox{0.67}{$\bm{MVBench}$}} & \rotatebox{65}{\scalebox{0.67}{$\bm{MSVD}$}} \\
Cap$\xrightarrow{}$All        & 1695.5 & 87.5    & 77.2    & 50.7 & 63.8    & 72.5 \\
Seg$\xrightarrow{}$Det$\xrightarrow{}$Cls$\xrightarrow{}$Cap        & 1627.2 & 85.5  & 74.9    & 48.3 & 61.3    & 70.6 \\
Cap$\xrightarrow{}$Cls$\xrightarrow{}$Det$\xrightarrow{}$Seg        & \textbf{1749.5} & \textbf{88.2}  & \textbf{78.2}    & \textbf{51.4} & \textbf{64.2}    & \textbf{73.3} \\
\end{tblr}
}
\caption{Ablation study on the order of progressive pre-alignment.}
\label{table:6}
\end{table}

\subsection{Ablation Study}
As shown in Fig. \ref{fig:3}, each metric in every stage shows an upward trend as the number of training steps increases. Each time the model reaches a stopping point in optimization and switches to the next stage, new gains are achieved, indicating that the training strategy is effective. As shown in Tab. \ref{table:3}, as the number of experts involved in decision-making increases from 1 to 3, the model’s performance gradually improves across tasks. With k=3, the model achieves peak performance across multiple metrics, while a slight decrease is observed at k=4, indicating that selecting three experts for decision-making is optimal. This phenomenon suggests that moderately increasing the number of experts helps improve the model’s overall performance, but too many experts may introduce redundant information or excessive complexity, thereby affecting performance. As shown in Tab. \ref{table:4}, by utilizing contrastive learning and residual connection, our method achieved better results compared to not utilizing the architecture, thus demonstrating the function and necessity of these two structures in our method. As shown in Tab. \ref{table:5}, in the Caption stage, we tested OpenAI CLIP ViT-L/14@336 \cite{radford2021learning}, EVA-CLIP-02 ViT-L/14@336 \cite{tong2024cambrian}, DINOv2 ViT-L/14@336 \cite{oquab2023dinov2}, and SigLIP ViT-SO400M/14@384 \cite{yin2024sea} as feature extractors. In the Classification stage, we tested RAM++ \cite{huang2023inject} and RAM as feature extractors. For the Detection stage, we evaluated Grounding-DINO-1.5 and Grounding-DINO-1.0 as feature extractors, and in the Segmentation stage, we tested SAM and SAM2 as feature extractors. Ultimately, the combination of SigLIP ViT-SO400M/14@384, RAM++, Grounding-DINO-1.5, and SAM2 yields the best results.
This is because models with higher performance generally yield better training results. Additionally, Dinov2, as a self-supervised visual backbone trained across multiple visual dimensions, excels in visual tasks. As shown in Tab. \ref{table:6}, following our progressive training strategy can achieve the best results on all metrics, and changing the default order or using a synchronous approach will result in a decrease in the performance of the model on all datasets.

%% file: sec/5_conclusion.tex
\section{Conclusion}
In this paper, we present the Astrea, introducing a progressive pre-alignment strategy that reduces the training load on the main model,while innovative training techniques help mitigate knowledge forgetting. Additionally, Astrea addresses the challenge of trade-off between multi-task heterogeneity and model generalization. The model is further enhanced by employing the MoE architecture, dynamic feature fusion, and momentum contrast learning.Astrea provides a robust and scalable solution for large-scale visual language modeling with heterogeneous expert collaboration and knowledge fusion strategies, setting a new standard for visual comprehension tasks. Our work makes a contribution in the field of multimodal large language models.

%% file: sec/X_suppl.tex
\clearpage
\setcounter{page}{1}
\setcounter{section}{0}
\section{Why From Coarse To Fine?}\label{supply}
As shown in Fig. \ref{fig:1}, the pre-alignment involves four visual tasks, from coarse to fine, namely Caption, Classification, Detection, Segment. 

The Caption task mainly focuses on the overall understanding and description of the image, using high-level visual information to describe the image, making it the most coarse-grain task.

The Classification task focuses on the features of the object rather than the image as a whole, and its special focus is also mainly on the high-level features of the object, so its grain size is also coarser, but finer than the caption task.

The Detection task is an instance-level, medium-grained analysis that requires localization and identification of individual objects (bounding boxes + categories), balancing local features with spatial relationships. 

The Segment task is pixel-level fine-grained parsing, which assigns semantic labels to each pixel and forces the model to understand the topology of the scene. Therefore, this type of task has the finest granularity.

Based on the above, we have designed a series of pre-alignment strategies from coarse to fine.

\begin{figure*}[t]
  \centering
  \includegraphics[width=\linewidth]{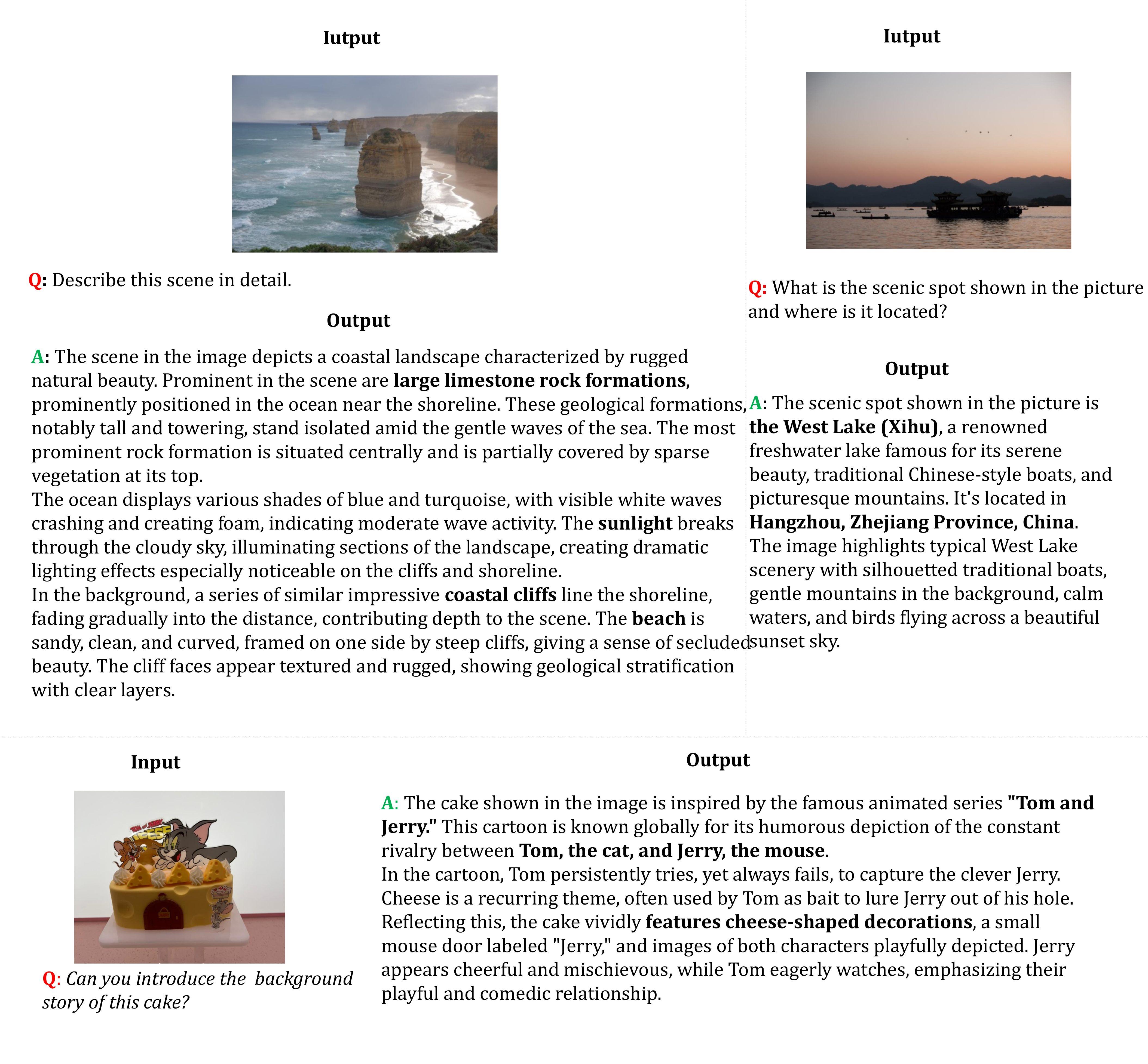}
  \caption{Demo of \textbf{Image Description} (left), \textbf{Landmark Recognition} (right) and \textbf{General Knowledge} (bottom).}
  \label{fig:4}
\end{figure*}

\begin{figure*}[t]
  \centering
  \includegraphics[width=\linewidth]{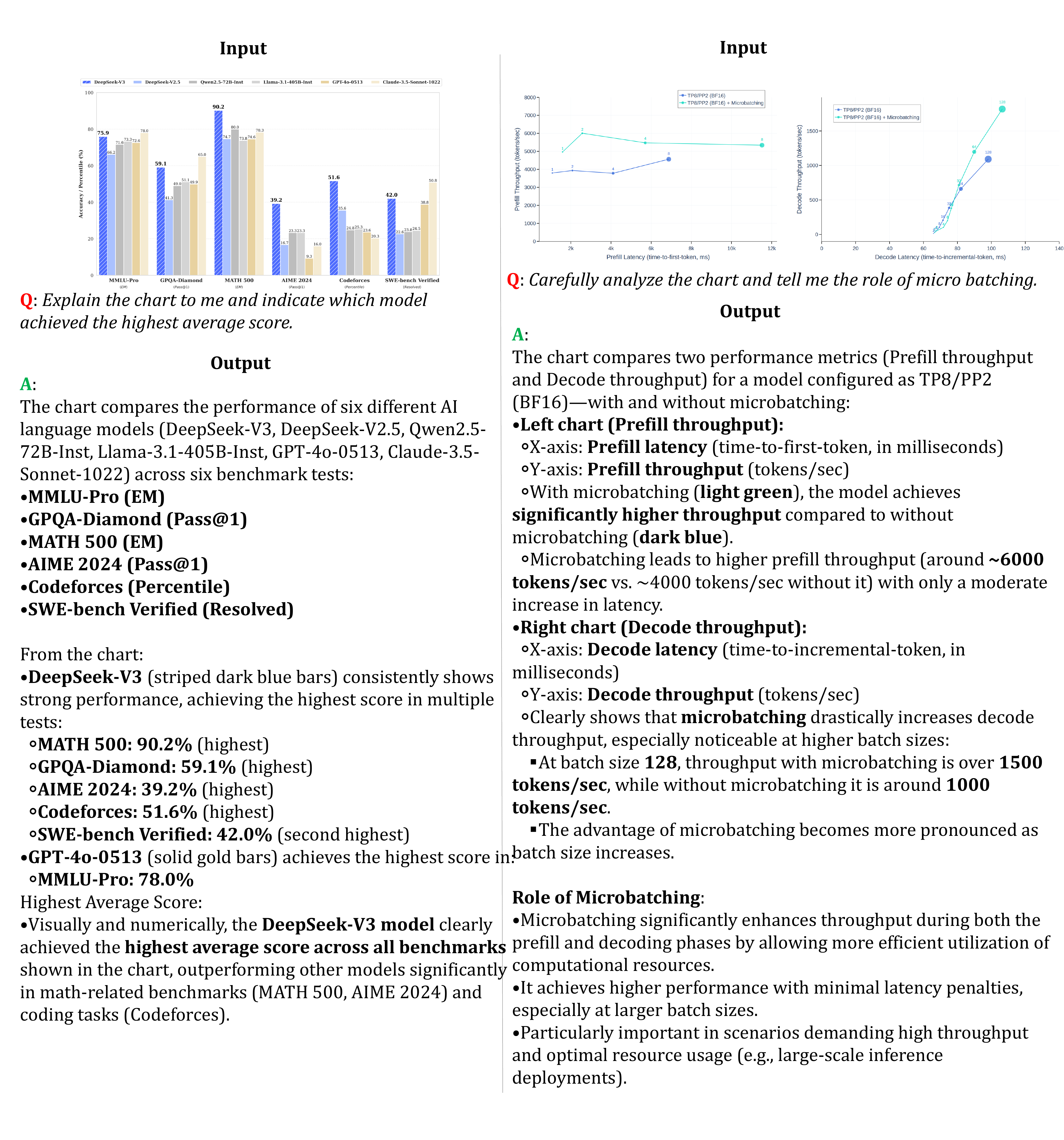}
  \caption{Demo of \textbf{Chart Understanding}.}
  \label{fig:5}
\end{figure*}

\begin{figure*}[t]
  \centering
  \includegraphics[width=\linewidth]{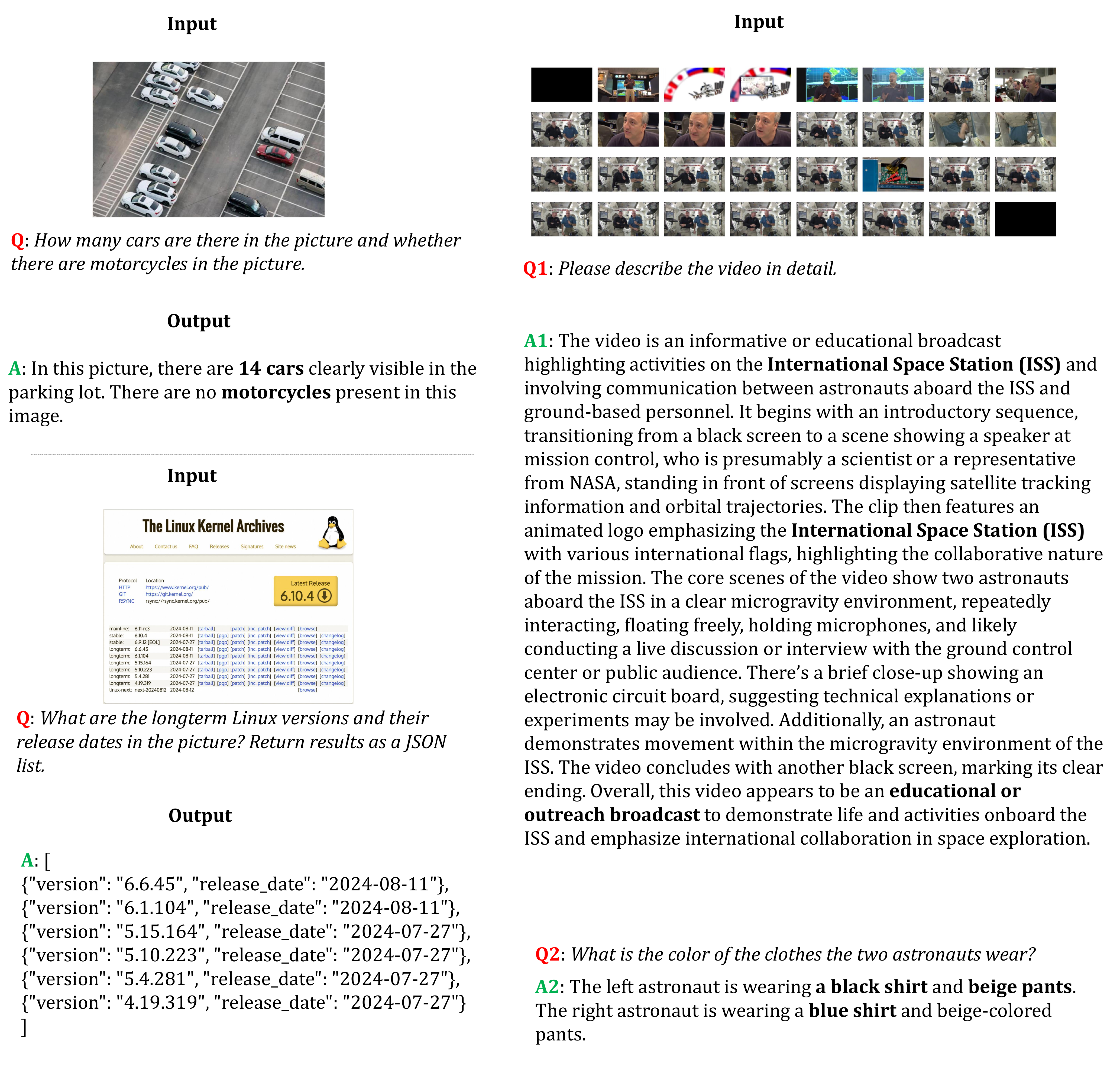}
  \caption{Demo of \textbf{Counting} (upper left), \textbf{ OCR and Format Following} (bottom left) and \textbf{Multi-Round Video Chat} (right).}
  \label{fig:6}
\end{figure*}

\section{Samples}
In the main text, we present a set of demo examples. We present more demo examples here to demonstrate the powerful performance of our model from different capabilities.

The left panel of Fig. \ref{fig:4} shows the excellent performance of Astrea in \textbf{Image Description} tasks, capturing background and environmental elements in the image very well. And the right panel of Fig. \ref{fig:4} presents Astrea's powerful analytical ability in \textbf{Landscape Recognition} tasks, and accurately analyzes the results based on landscape features. In the bottom panel of Fig. \ref{fig:4}, Astrea not only mentions the main characters of the animation - Tom Cat and Jerry Mouse, but also explains their relationship and the importance of cheese in the animation, helping readers better understand the \textbf{General Knowledge}.
Fig. \ref{fig:5} demonstrates the ability of Astrea in \textbf{Chart Understanding}, which not only provides accurate, comprehensive, and easy to understand data analysis, but also analyzes the role of specific elements in the chart. The left part of Fig. \ref{fig:6} shows the superiority of Astrea in \textbf{counting} and \textbf{OCR} capabilities, while the right part of Fig. \ref{fig:6} emphasizes our Astrea's ability to understand and analyze video details in \textbf{Multi-Round Video Chat}, as well as its ability to accurately output specific component features.

\section{Other Related Work}
\subsection{Residual Connection}
A residual connection is a mechanism used to address the issue of vanishing or exploding gradients that can occur as the depth of a neural network increases. It was first introduced in \cite{he2016deep}, with the core idea of adding the residual back to the output of deeper layers. This reduces the degradation of gradient information as it flows backward through multiple layers during training. Residual connections allow the network to focus on learning incremental adjustments instead of trying to learn the entire transformation from the beginning.
Residual connections have been widely adopted by various models, ranging from the original ResNet in computer vision \cite{he2016deep}, to Transformer in natural language processing \cite{vaswani2017attention}, and extending to AlphaZero in reinforcement learning \cite{silver2018general} and AlphaFold for protein structure prediction \cite{senior2020improved}.

\subsection{Contrastive Learning}
Contrast learning, as an important paradigm of self-supervised learning, has made significant progress in the field of feature representation learning in recent years. Momentum Contrast Learning (MoCo) \cite{he2020momentum} effectively solves the contradiction between the restricted negative sample capacity and feature consistency in traditional contrast learning through the dynamic dictionary cohort construction and momentum encoder update mechanism. The method adopts a queue-type negative sample storage strategy, which enables the model to construct a large-scale contrast dictionary using historical batch samples, and at the same time controls the parameter update speed of the key-value encoder through the momentum coefficients to ensure the stability of the feature space. This type of method provides theoretical support for cross-modal representation alignment, and its proposed temperature coefficient adjustment mechanism is also widely used to alleviate the problem of multimodal data distribution differences, and these technical accumulations lay an important foundation for the design of the feature fusion framework in this study.

\section{Dataset and Instruct}\label{dataset}
For image understanding tasks, training datasets can be categorized into several types: OCR datasets are primarily used for optical character recognition tasks and include examples such as Filtered DVQA, DVQA \cite{kafle2018dvqa}, and SynthDog. General datasets are used for general dialogue and visual understanding tasks, such as ALLaVA \cite{chen2024allava}, Q-Instruct \cite{wu2024q}, and DocVQA \cite{mathew2021docvqa}. Language datasets focus on language-related tasks and include OpenOrca \cite{lian2023openorca} and MathInstruct \cite{yue2023mammoth}. Counting datasets are dedicated to object counting tasks, such as Filtered CLEVR \cite{johnson2017inferring} and TallyQA \cite{acharya2019tallyqa}. Code datasets cover programming dialogues, such as Design2Code \cite{si2024design2code}. Math datasets are for mathematical tasks and include Geo170K and RAVEN. Science datasets pertain to scientific knowledge and visual question-answering tasks, such as PathVQA \cite{he2020pathvqa} and ScienceQA \cite{lu2022learn}. We conducted a comprehensive evaluation of the model on representative datasets across different task categories. The test set covers categories such as general question-answering, knowledge domains, OCR and chart analysis, as well as vision-centric tasks.

For video understanding tasks, we leveraged a teaching video-to-text dataset to extend the capability of Multimodal Large Language Models (MLLMs) to process video inputs. The training data includes a variety of datasets for video understanding tasks: 27k dialogue videos from VideoChat \cite{li2023videochat} and Video-ChatGPT \cite{maaz2023video}; 80k classification task samples from Kinetics \cite{kay2017kinetics} and SthSthV2 \cite{goyal2017something}; 450k captioned video samples from Webvid \cite{bain2021frozen}, YouCook2 \cite{zhou2018towards}, TextVR \cite{wu2025large}, and VideoChat; 117k reasoning data samples from NextQA \cite{xiao2021next} and CLEVRER \cite{yi2019clevrer}; and 109k annotated question-answering samples from Webvid, TGIF \cite{li2016tgif}, and Ego4D \cite{grauman2022ego4d}. Overall, we used 783k instructional fine-tuning data. We evaluated the trained model using the following video-to-text benchmark tests. First, the open-ended Video Question Answering (VideoQA) benchmarks include MSVD-QA \cite{xu2017video}, MSRVTT-QA \cite{xu2017video}, ActivityQA \cite{yu2019activitynet}, and TGIF QA \cite{li2016tgif}. The answers in these benchmarks are typically composed of single words.

Different datasets are provided with various prompts, including multiple-choice, true/false, short answer, caption, and location-based questions. For example, prompts might include questions like ‘What brand of watch is this?’, ‘Is there a car in the image?’, ‘What has the yellow object been drawn to resemble? Choose from: eagle, face, dog, star’.

%% file: main.bbl
\begin{thebibliography}{72}
\providecommand{\natexlab}[1]{#1}
\providecommand{\url}[1]{\texttt{#1}}
\expandafter\ifx\csname urlstyle\endcsname\relax
  \providecommand{\doi}[1]{doi: #1}\else
  \providecommand{\doi}{doi: \begingroup \urlstyle{rm}\Url}\fi

\bibitem[Acharya et~al.(2019)Acharya, Kafle, and Kanan]{acharya2019tallyqa}
Manoj Acharya, Kushal Kafle, and Christopher Kanan.
\newblock Tallyqa: Answering complex counting questions.
\newblock In \emph{Proceedings of the AAAI conference on artificial intelligence}, pages 8076--8084, 2019.

\bibitem[Anderson et~al.(2018)Anderson, He, Buehler, Teney, Johnson, Gould, and Zhang]{anderson2018bottom}
Peter Anderson, Xiaodong He, Chris Buehler, Damien Teney, Mark Johnson, Stephen Gould, and Lei Zhang.
\newblock Bottom-up and top-down attention for image captioning and visual question answering.
\newblock In \emph{Proceedings of the IEEE conference on computer vision and pattern recognition}, pages 6077--6086, 2018.

\bibitem[Bain et~al.(2021)Bain, Nagrani, Varol, and Zisserman]{bain2021frozen}
Max Bain, Arsha Nagrani, G{\"u}l Varol, and Andrew Zisserman.
\newblock Frozen in time: A joint video and image encoder for end-to-end retrieval.
\newblock In \emph{Proceedings of the IEEE/CVF international conference on computer vision}, pages 1728--1738, 2021.

\bibitem[Cai et~al.(2024)Cai, Jiang, Wang, Tang, Kim, and Huang]{cai2024survey}
Weilin Cai, Juyong Jiang, Fan Wang, Jing Tang, Sunghun Kim, and Jiayi Huang.
\newblock A survey on mixture of experts.
\newblock \emph{arXiv preprint arXiv:2407.06204}, 2024.

\bibitem[Chang et~al.(2024)Chang, Wang, Wang, Wu, Yang, Zhu, Chen, Yi, Wang, Wang, et~al.]{chang2024survey}
Yupeng Chang, Xu Wang, Jindong Wang, Yuan Wu, Linyi Yang, Kaijie Zhu, Hao Chen, Xiaoyuan Yi, Cunxiang Wang, Yidong Wang, et~al.
\newblock A survey on evaluation of large language models.
\newblock \emph{ACM Transactions on Intelligent Systems and Technology}, 15\penalty0 (3):\penalty0 1--45, 2024.

\bibitem[Chen et~al.(2024)Chen, Chen, Zhang, Chen, Wu, Zhang, Chen, Li, Wan, and Wang]{chen2024allava}
Guiming~Hardy Chen, Shunian Chen, Ruifei Zhang, Junying Chen, Xiangbo Wu, Zhiyi Zhang, Zhihong Chen, Jianquan Li, Xiang Wan, and Benyou Wang.
\newblock Allava: Harnessing gpt4v-synthesized data for a lite vision-language model.
\newblock \emph{arXiv preprint arXiv:2402.11684}, 2024.

\bibitem[Chen et~al.(2020{\natexlab{a}})Chen, Kornblith, Norouzi, and Hinton]{chen2020simple}
Ting Chen, Simon Kornblith, Mohammad Norouzi, and Geoffrey Hinton.
\newblock A simple framework for contrastive learning of visual representations.
\newblock In \emph{International conference on machine learning}, pages 1597--1607. PMLR, 2020{\natexlab{a}}.

\bibitem[Chen et~al.(2020{\natexlab{b}})Chen, Li, Yu, El~Kholy, Ahmed, Gan, Cheng, and Liu]{chen2020uniter}
Yen-Chun Chen, Linjie Li, Licheng Yu, Ahmed El~Kholy, Faisal Ahmed, Zhe Gan, Yu Cheng, and Jingjing Liu.
\newblock Uniter: Universal image-text representation learning.
\newblock In \emph{European conference on computer vision}, pages 104--120. Springer, 2020{\natexlab{b}}.

\bibitem[Deng et~al.(2009)Deng, Dong, Socher, Li, Li, and Fei-Fei]{deng2009imagenet}
Jia Deng, Wei Dong, Richard Socher, Li-Jia Li, Kai Li, and Li Fei-Fei.
\newblock Imagenet: A large-scale hierarchical image database.
\newblock In \emph{2009 IEEE conference on computer vision and pattern recognition}, pages 248--255. Ieee, 2009.

\bibitem[Dosovitskiy(2020)]{dosovitskiy2020image}
Alexey Dosovitskiy.
\newblock An image is worth 16x16 words: Transformers for image recognition at scale.
\newblock \emph{arXiv preprint arXiv:2010.11929}, 2020.

\bibitem[Dou et~al.(2023)Dou, Zhou, Liu, Gao, Zhao, Shen, Zhou, Xi, Wang, Fan, Pu, Zhu, Zheng, Gui, Zhang, and Huang]{dou2024loramoe}
Shihan Dou, Enyu Zhou, Yan Liu, Songyang Gao, Jun Zhao, Wei Shen, Yuhao Zhou, Zhiheng Xi, Xiao Wang, Xiaoran Fan, Shiliang Pu, Jiang Zhu, Rui Zheng, Tao Gui, Qi Zhang, and Xuanjing Huang.
\newblock Loramoe: Revolutionizing mixture of experts for maintaining world knowledge in language model alignment, 2023.

\bibitem[Fang et~al.(2023)Fang, Jose, Jain, Schmidt, Toshev, and Shankar]{fang2023data}
Alex Fang, Albin~Madappally Jose, Amit Jain, Ludwig Schmidt, Alexander Toshev, and Vaishaal Shankar.
\newblock Data filtering networks.
\newblock \emph{arXiv preprint arXiv:2309.17425}, 2023.

\bibitem[Fedus et~al.(2022)Fedus, Zoph, and Shazeer]{fedus2022switch}
William Fedus, Barret Zoph, and Noam Shazeer.
\newblock Switch transformers: Scaling to trillion parameter models with simple and efficient sparsity.
\newblock \emph{Journal of Machine Learning Research}, 23\penalty0 (120):\penalty0 1--39, 2022.

\bibitem[Ge et~al.(2023)Ge, Ge, Zeng, Wang, and Shan]{ge2023planting}
Yuying Ge, Yixiao Ge, Ziyun Zeng, Xintao Wang, and Ying Shan.
\newblock Planting a seed of vision in large language model.
\newblock \emph{arXiv preprint arXiv:2307.08041}, 2023.

\bibitem[Goyal et~al.(2017)Goyal, Ebrahimi~Kahou, Michalski, Materzynska, Westphal, Kim, Haenel, Fruend, Yianilos, Mueller-Freitag, et~al.]{goyal2017something}
Raghav Goyal, Samira Ebrahimi~Kahou, Vincent Michalski, Joanna Materzynska, Susanne Westphal, Heuna Kim, Valentin Haenel, Ingo Fruend, Peter Yianilos, Moritz Mueller-Freitag, et~al.
\newblock The" something something" video database for learning and evaluating visual common sense.
\newblock In \emph{Proceedings of the IEEE international conference on computer vision}, pages 5842--5850, 2017.

\bibitem[Grauman et~al.(2022)Grauman, Westbury, Byrne, Chavis, Furnari, Girdhar, Hamburger, Jiang, Liu, Liu, et~al.]{grauman2022ego4d}
Kristen Grauman, Andrew Westbury, Eugene Byrne, Zachary Chavis, Antonino Furnari, Rohit Girdhar, Jackson Hamburger, Hao Jiang, Miao Liu, Xingyu Liu, et~al.
\newblock Ego4d: Around the world in 3,000 hours of egocentric video.
\newblock In \emph{Proceedings of the IEEE/CVF Conference on Computer Vision and Pattern Recognition}, pages 18995--19012, 2022.

\bibitem[He et~al.(2016)He, Zhang, Ren, and Sun]{he2016deep}
Kaiming He, Xiangyu Zhang, Shaoqing Ren, and Jian Sun.
\newblock Deep residual learning for image recognition.
\newblock In \emph{Proceedings of the IEEE conference on computer vision and pattern recognition}, pages 770--778, 2016.

\bibitem[He et~al.(2020{\natexlab{a}})He, Fan, Wu, Xie, and Girshick]{he2020momentum}
Kaiming He, Haoqi Fan, Yuxin Wu, Saining Xie, and Ross Girshick.
\newblock Momentum contrast for unsupervised visual representation learning.
\newblock In \emph{Proceedings of the IEEE/CVF conference on computer vision and pattern recognition}, pages 9729--9738, 2020{\natexlab{a}}.

\bibitem[He et~al.(2020{\natexlab{b}})He, Zhang, Mou, Xing, and Xie]{he2020pathvqa}
Xuehai He, Yichen Zhang, Luntian Mou, Eric Xing, and Pengtao Xie.
\newblock Pathvqa: 30000+ questions for medical visual question answering.
\newblock \emph{arXiv preprint arXiv:2003.10286}, 2020{\natexlab{b}}.

\bibitem[Hiippala et~al.(2021)Hiippala, Alikhani, Haverinen, Kalliokoski, Logacheva, Orekhova, Tuomainen, Stone, and Bateman]{hiippala2021ai2d}
Tuomo Hiippala, Malihe Alikhani, Jonas Haverinen, Timo Kalliokoski, Evanfiya Logacheva, Serafina Orekhova, Aino Tuomainen, Matthew Stone, and John~A Bateman.
\newblock Ai2d-rst: A multimodal corpus of 1000 primary school science diagrams.
\newblock \emph{Language Resources and Evaluation}, 55:\penalty0 661--688, 2021.

\bibitem[Huang et~al.(2023)Huang, Huang, Zhang, Tian, Feng, Zhang, Xie, Li, and Zhang]{huang2023inject}
Xinyu Huang, Yi-Jie Huang, Youcai Zhang, Weiwei Tian, Rui Feng, Yuejie Zhang, Yanchun Xie, Yaqian Li, and Lei Zhang.
\newblock Inject semantic concepts into image tagging for open-set recognition.
\newblock \emph{arXiv preprint arXiv:2310.15200}, 2023.

\bibitem[Hudson and Manning(2019)]{hudson2019gqa}
Drew~A Hudson and Christopher~D Manning.
\newblock Gqa: A new dataset for real-world visual reasoning and compositional question answering.
\newblock In \emph{Proceedings of the IEEE/CVF conference on computer vision and pattern recognition}, pages 6700--6709, 2019.

\bibitem[Johnson et~al.(2017)Johnson, Hariharan, Van Der~Maaten, Hoffman, Fei-Fei, Lawrence~Zitnick, and Girshick]{johnson2017inferring}
Justin Johnson, Bharath Hariharan, Laurens Van Der~Maaten, Judy Hoffman, Li Fei-Fei, C Lawrence~Zitnick, and Ross Girshick.
\newblock Inferring and executing programs for visual reasoning.
\newblock In \emph{Proceedings of the IEEE international conference on computer vision}, pages 2989--2998, 2017.

\bibitem[Kafle et~al.(2018)Kafle, Price, Cohen, and Kanan]{kafle2018dvqa}
Kushal Kafle, Brian Price, Scott Cohen, and Christopher Kanan.
\newblock Dvqa: Understanding data visualizations via question answering.
\newblock In \emph{Proceedings of the IEEE conference on computer vision and pattern recognition}, pages 5648--5656, 2018.

\bibitem[Kay et~al.(2017)Kay, Carreira, Simonyan, Zhang, Hillier, Vijayanarasimhan, Viola, Green, Back, Natsev, et~al.]{kay2017kinetics}
Will Kay, Joao Carreira, Karen Simonyan, Brian Zhang, Chloe Hillier, Sudheendra Vijayanarasimhan, Fabio Viola, Tim Green, Trevor Back, Paul Natsev, et~al.
\newblock The kinetics human action video dataset.
\newblock \emph{arXiv preprint arXiv:1705.06950}, 2017.

\bibitem[Krizhevsky et~al.(2012)Krizhevsky, Sutskever, and Hinton]{krizhevsky2012imagenet}
Alex Krizhevsky, Ilya Sutskever, and Geoffrey~E Hinton.
\newblock Imagenet classification with deep convolutional neural networks.
\newblock \emph{Advances in neural information processing systems}, 25, 2012.

\bibitem[LeCun et~al.(2015)LeCun, Bengio, and Hinton]{lecun2015deep}
Yann LeCun, Yoshua Bengio, and Geoffrey Hinton.
\newblock Deep learning.
\newblock \emph{nature}, 521\penalty0 (7553):\penalty0 436--444, 2015.

\bibitem[Li et~al.(2024)Li, Zhang, Guo, Zhang, Li, Zhang, Zhang, Li, Liu, and Li]{li2024llava}
Bo Li, Yuanhan Zhang, Dong Guo, Renrui Zhang, Feng Li, Hao Zhang, Kaichen Zhang, Yanwei Li, Ziwei Liu, and Chunyuan Li.
\newblock Llava-onevision: Easy visual task transfer.
\newblock \emph{arXiv preprint arXiv:2408.03326}, 2024.

\bibitem[Li et~al.(2022)Li, Xu, Tian, Wang, Yan, Bi, Ye, Chen, Xu, Cao, et~al.]{li2022mplug}
Chenliang Li, Haiyang Xu, Junfeng Tian, Wei Wang, Ming Yan, Bin Bi, Jiabo Ye, Hehong Chen, Guohai Xu, Zheng Cao, et~al.
\newblock mplug: Effective and efficient vision-language learning by cross-modal skip-connections.
\newblock \emph{arXiv preprint arXiv:2205.12005}, 2022.

\bibitem[Li et~al.(2023)Li, He, Wang, Li, Wang, Luo, Wang, Wang, and Qiao]{li2023videochat}
KunChang Li, Yinan He, Yi Wang, Yizhuo Li, Wenhai Wang, Ping Luo, Yali Wang, Limin Wang, and Yu Qiao.
\newblock Videochat: Chat-centric video understanding.
\newblock \emph{arXiv preprint arXiv:2305.06355}, 2023.

\bibitem[Li et~al.(2020)Li, Yin, Li, Zhang, Hu, Zhang, Wang, Hu, Dong, Wei, et~al.]{li2020oscar}
Xiujun Li, Xi Yin, Chunyuan Li, Pengchuan Zhang, Xiaowei Hu, Lei Zhang, Lijuan Wang, Houdong Hu, Li Dong, Furu Wei, et~al.
\newblock Oscar: Object-semantics aligned pre-training for vision-language tasks.
\newblock In \emph{Computer Vision--ECCV 2020: 16th European Conference, Glasgow, UK, August 23--28, 2020, Proceedings, Part XXX 16}, pages 121--137. Springer, 2020.

\bibitem[Li et~al.(2016)Li, Song, Cao, Tetreault, Goldberg, Jaimes, and Luo]{li2016tgif}
Yuncheng Li, Yale Song, Liangliang Cao, Joel Tetreault, Larry Goldberg, Alejandro Jaimes, and Jiebo Luo.
\newblock Tgif: A new dataset and benchmark on animated gif description.
\newblock In \emph{Proceedings of the IEEE Conference on Computer Vision and Pattern Recognition}, pages 4641--4650, 2016.

\bibitem[Lian et~al.(2023)Lian, Goodson, Pentland, et~al.]{lian2023openorca}
W Lian, B Goodson, E Pentland, et~al.
\newblock Openorca: An open dataset of gpt augmented flan reasoning traces, 2023.

\bibitem[Lin et~al.(2024)Lin, Tang, Ye, Cui, Zhu, Jin, Huang, Zhang, Pang, Ning, et~al.]{lin2024moe}
Bin Lin, Zhenyu Tang, Yang Ye, Jiaxi Cui, Bin Zhu, Peng Jin, Jinfa Huang, Junwu Zhang, Yatian Pang, Munan Ning, et~al.
\newblock Moe-llava: Mixture of experts for large vision-language models.
\newblock \emph{arXiv preprint arXiv:2401.15947}, 2024.

\bibitem[Lin et~al.(2014)Lin, Maire, Belongie, Hays, Perona, Ramanan, Doll{\'a}r, and Zitnick]{lin2014microsoft}
Tsung-Yi Lin, Michael Maire, Serge Belongie, James Hays, Pietro Perona, Deva Ramanan, Piotr Doll{\'a}r, and C~Lawrence Zitnick.
\newblock Microsoft coco: Common objects in context.
\newblock In \emph{Computer Vision--ECCV 2014: 13th European Conference, Zurich, Switzerland, September 6-12, 2014, Proceedings, Part V 13}, pages 740--755. Springer, 2014.

\bibitem[Liu et~al.(2024)Liu, Li, Li, Li, Zhang, Shen, and Lee]{liu2024llava}
Haotian Liu, Chunyuan Li, Yuheng Li, Bo Li, Yuanhan Zhang, Sheng Shen, and Yong~Jae Lee.
\newblock Llava-next: Improved reasoning, ocr, and world knowledge, 2024.

\bibitem[Liu et~al.(2023)Liu, Li, Yang, Li, Yin, Liu, Jin, and Bai]{liu2023hidden}
Yuliang Liu, Zhang Li, Biao Yang, Chunyuan Li, Xucheng Yin, Cheng-lin Liu, Lianwen Jin, and Xiang Bai.
\newblock On the hidden mystery of ocr in large multimodal models.
\newblock \emph{arXiv preprint arXiv:2305.07895}, 2023.

\bibitem[Liu et~al.(2025)Liu, Duan, Zhang, Li, Zhang, Zhao, Yuan, Wang, He, Liu, et~al.]{liu2025mmbench}
Yuan Liu, Haodong Duan, Yuanhan Zhang, Bo Li, Songyang Zhang, Wangbo Zhao, Yike Yuan, Jiaqi Wang, Conghui He, Ziwei Liu, et~al.
\newblock Mmbench: Is your multi-modal model an all-around player?
\newblock In \emph{European Conference on Computer Vision}, pages 216--233. Springer, 2025.

\bibitem[Lu et~al.(2022)Lu, Mishra, Xia, Qiu, Chang, Zhu, Tafjord, Clark, and Kalyan]{lu2022learn}
Pan Lu, Swaroop Mishra, Tanglin Xia, Liang Qiu, Kai-Wei Chang, Song-Chun Zhu, Oyvind Tafjord, Peter Clark, and Ashwin Kalyan.
\newblock Learn to explain: Multimodal reasoning via thought chains for science question answering.
\newblock \emph{Advances in Neural Information Processing Systems}, 35:\penalty0 2507--2521, 2022.

\bibitem[Lu et~al.(2023)Lu, Bansal, Xia, Liu, Li, Hajishirzi, Cheng, Chang, Galley, and Gao]{lu2023mathvista}
Pan Lu, Hritik Bansal, Tony Xia, Jiacheng Liu, Chunyuan Li, Hannaneh Hajishirzi, Hao Cheng, Kai-Wei Chang, Michel Galley, and Jianfeng Gao.
\newblock Mathvista: Evaluating mathematical reasoning of foundation models in visual contexts.
\newblock \emph{arXiv preprint arXiv:2310.02255}, 2023.

\bibitem[Maaz et~al.(2023)Maaz, Rasheed, Khan, and Khan]{maaz2023video}
Muhammad Maaz, Hanoona Rasheed, Salman Khan, and Fahad~Shahbaz Khan.
\newblock Video-chatgpt: Towards detailed video understanding via large vision and language models.
\newblock \emph{arXiv preprint arXiv:2306.05424}, 2023.

\bibitem[Masry et~al.(2022)Masry, Long, Tan, Joty, and Hoque]{masry2022chartqa}
Ahmed Masry, Do~Xuan Long, Jia~Qing Tan, Shafiq Joty, and Enamul Hoque.
\newblock Chartqa: A benchmark for question answering about charts with visual and logical reasoning.
\newblock \emph{arXiv preprint arXiv:2203.10244}, 2022.

\bibitem[Mathew et~al.(2021)Mathew, Karatzas, and Jawahar]{mathew2021docvqa}
Minesh Mathew, Dimosthenis Karatzas, and CV Jawahar.
\newblock Docvqa: A dataset for vqa on document images.
\newblock In \emph{Proceedings of the IEEE/CVF winter conference on applications of computer vision}, pages 2200--2209, 2021.

\bibitem[Mnih et~al.(2014)Mnih, Heess, Graves, et~al.]{mnih2014recurrent}
Volodymyr Mnih, Nicolas Heess, Alex Graves, et~al.
\newblock Recurrent models of visual attention.
\newblock \emph{Advances in neural information processing systems}, 27, 2014.

\bibitem[Noroozi and Favaro(2016)]{noroozi2016unsupervised}
Mehdi Noroozi and Paolo Favaro.
\newblock Unsupervised learning of visual representations by solving jigsaw puzzles.
\newblock In \emph{European conference on computer vision}, pages 69--84. Springer, 2016.

\bibitem[Oquab et~al.(2023)Oquab, Darcet, Moutakanni, Vo, Szafraniec, Khalidov, Fernandez, Haziza, Massa, El-Nouby, et~al.]{oquab2023dinov2}
Maxime Oquab, Timoth{\'e}e Darcet, Th{\'e}o Moutakanni, Huy Vo, Marc Szafraniec, Vasil Khalidov, Pierre Fernandez, Daniel Haziza, Francisco Massa, Alaaeldin El-Nouby, et~al.
\newblock Dinov2: Learning robust visual features without supervision.
\newblock \emph{arXiv preprint arXiv:2304.07193}, 2023.

\bibitem[Radford et~al.(2021)Radford, Kim, Hallacy, Ramesh, Goh, Agarwal, Sastry, Askell, Mishkin, Clark, et~al.]{radford2021learning}
Alec Radford, Jong~Wook Kim, Chris Hallacy, Aditya Ramesh, Gabriel Goh, Sandhini Agarwal, Girish Sastry, Amanda Askell, Pamela Mishkin, Jack Clark, et~al.
\newblock Learning transferable visual models from natural language supervision.
\newblock In \emph{International conference on machine learning}, pages 8748--8763. PMLR, 2021.

\bibitem[Redmon(2016)]{redmon2016you}
J Redmon.
\newblock You only look once: Unified, real-time object detection.
\newblock In \emph{Proceedings of the IEEE conference on computer vision and pattern recognition}, 2016.

\bibitem[Ren et~al.(2016)Ren, He, Girshick, and Sun]{ren2016faster}
Shaoqing Ren, Kaiming He, Ross Girshick, and Jian Sun.
\newblock Faster r-cnn: Towards real-time object detection with region proposal networks.
\newblock \emph{IEEE transactions on pattern analysis and machine intelligence}, 39\penalty0 (6):\penalty0 1137--1149, 2016.

\bibitem[Senior et~al.(2020)Senior, Evans, Jumper, Kirkpatrick, Sifre, Green, Qin, {\v{Z}}{\'\i}dek, Nelson, Bridgland, et~al.]{senior2020improved}
Andrew~W Senior, Richard Evans, John Jumper, James Kirkpatrick, Laurent Sifre, Tim Green, Chongli Qin, Augustin {\v{Z}}{\'\i}dek, Alexander~WR Nelson, Alex Bridgland, et~al.
\newblock Improved protein structure prediction using potentials from deep learning.
\newblock \emph{Nature}, 577\penalty0 (7792):\penalty0 706--710, 2020.

\bibitem[Si et~al.(2024)Si, Zhang, Yang, Liu, and Yang]{si2024design2code}
Chenglei Si, Yanzhe Zhang, Zhengyuan Yang, Ruibo Liu, and Diyi Yang.
\newblock Design2code: How far are we from automating front-end engineering?
\newblock \emph{arXiv preprint arXiv:2403.03163}, 2024.

\bibitem[Silver et~al.(2018)Silver, Hubert, Schrittwieser, Antonoglou, Lai, Guez, Lanctot, Sifre, Kumaran, Graepel, et~al.]{silver2018general}
David Silver, Thomas Hubert, Julian Schrittwieser, Ioannis Antonoglou, Matthew Lai, Arthur Guez, Marc Lanctot, Laurent Sifre, Dharshan Kumaran, Thore Graepel, et~al.
\newblock A general reinforcement learning algorithm that masters chess, shogi, and go through self-play.
\newblock \emph{Science}, 362\penalty0 (6419):\penalty0 1140--1144, 2018.

\bibitem[Singh et~al.(2019)Singh, Natarajan, Shah, Jiang, Chen, Batra, Parikh, and Rohrbach]{singh2019towards}
Amanpreet Singh, Vivek Natarajan, Meet Shah, Yu Jiang, Xinlei Chen, Dhruv Batra, Devi Parikh, and Marcus Rohrbach.
\newblock Towards vqa models that can read.
\newblock In \emph{Proceedings of the IEEE/CVF conference on computer vision and pattern recognition}, pages 8317--8326, 2019.

\bibitem[Su et~al.(2019)Su, Zhu, Cao, Li, Lu, Wei, and Dai]{su2019vl}
Weijie Su, Xizhou Zhu, Yue Cao, Bin Li, Lewei Lu, Furu Wei, and Jifeng Dai.
\newblock Vl-bert: Pre-training of generic visual-linguistic representations.
\newblock \emph{arXiv preprint arXiv:1908.08530}, 2019.

\bibitem[Tong et~al.(2024{\natexlab{a}})Tong, Brown, Wu, Woo, Middepogu, Akula, Yang, Yang, Iyer, Pan, et~al.]{tong2024cambrian}
Shengbang Tong, Ellis Brown, Penghao Wu, Sanghyun Woo, Manoj Middepogu, Sai~Charitha Akula, Jihan Yang, Shusheng Yang, Adithya Iyer, Xichen Pan, et~al.
\newblock Cambrian-1: A fully open, vision-centric exploration of multimodal llms.
\newblock \emph{arXiv preprint arXiv:2406.16860}, 2024{\natexlab{a}}.

\bibitem[Tong et~al.(2024{\natexlab{b}})Tong, Liu, Zhai, Ma, LeCun, and Xie]{tong2024eyes}
Shengbang Tong, Zhuang Liu, Yuexiang Zhai, Yi Ma, Yann LeCun, and Saining Xie.
\newblock Eyes wide shut? exploring the visual shortcomings of multimodal llms.
\newblock In \emph{Proceedings of the IEEE/CVF Conference on Computer Vision and Pattern Recognition}, pages 9568--9578, 2024{\natexlab{b}}.

\bibitem[Vaswani(2017)]{vaswani2017attention}
A Vaswani.
\newblock Attention is all you need.
\newblock \emph{Advances in Neural Information Processing Systems}, 2017.

\bibitem[Wang et~al.(2024)Wang, Bai, Tan, Wang, Fan, Bai, Chen, Liu, Wang, Ge, et~al.]{wang2024qwen2}
Peng Wang, Shuai Bai, Sinan Tan, Shijie Wang, Zhihao Fan, Jinze Bai, Keqin Chen, Xuejing Liu, Jialin Wang, Wenbin Ge, et~al.
\newblock Qwen2-vl: Enhancing vision-language model's perception of the world at any resolution.
\newblock \emph{arXiv preprint arXiv:2409.12191}, 2024.

\bibitem[Wu et~al.(2024)Wu, Zhang, Zhang, Chen, Liao, Wang, Xu, Li, Hou, Zhai, et~al.]{wu2024q}
Haoning Wu, Zicheng Zhang, Erli Zhang, Chaofeng Chen, Liang Liao, Annan Wang, Kaixin Xu, Chunyi Li, Jingwen Hou, Guangtao Zhai, et~al.
\newblock Q-instruct: Improving low-level visual abilities for multi-modality foundation models.
\newblock In \emph{Proceedings of the IEEE/CVF Conference on Computer Vision and Pattern Recognition}, pages 25490--25500, 2024.

\bibitem[Wu et~al.(2025)Wu, Zhao, Li, Li, Zhou, Shou, and Bai]{wu2025large}
Weijia Wu, Yuzhong Zhao, Zhuang Li, Jiahong Li, Hong Zhou, Mike~Zheng Shou, and Xiang Bai.
\newblock A large cross-modal video retrieval dataset with reading comprehension.
\newblock \emph{Pattern Recognition}, 157:\penalty0 110818, 2025.

\bibitem[Xiao et~al.(2021)Xiao, Shang, Yao, and Chua]{xiao2021next}
Junbin Xiao, Xindi Shang, Angela Yao, and Tat-Seng Chua.
\newblock Next-qa: Next phase of question-answering to explaining temporal actions.
\newblock In \emph{Proceedings of the IEEE/CVF conference on computer vision and pattern recognition}, pages 9777--9786, 2021.

\bibitem[Xu et~al.(2017)Xu, Zhao, Xiao, Wu, Zhang, He, and Zhuang]{xu2017video}
Dejing Xu, Zhou Zhao, Jun Xiao, Fei Wu, Hanwang Zhang, Xiangnan He, and Yueting Zhuang.
\newblock Video question answering via gradually refined attention over appearance and motion.
\newblock In \emph{Proceedings of the 25th ACM international conference on Multimedia}, pages 1645--1653, 2017.

\bibitem[Xu(2015)]{xu2015show}
Kelvin Xu.
\newblock Show, attend and tell: Neural image caption generation with visual attention.
\newblock \emph{arXiv preprint arXiv:1502.03044}, 2015.

\bibitem[Yi et~al.(2019)Yi, Gan, Li, Kohli, Wu, Torralba, and Tenenbaum]{yi2019clevrer}
Kexin Yi, Chuang Gan, Yunzhu Li, Pushmeet Kohli, Jiajun Wu, Antonio Torralba, and Joshua~B Tenenbaum.
\newblock Clevrer: Collision events for video representation and reasoning.
\newblock \emph{arXiv preprint arXiv:1910.01442}, 2019.

\bibitem[Yin et~al.(2024)Yin, Zhao, Zhang, Lin, Wang, Tao, Wan, Zhang, Yin, and Zhang]{yin2024sea}
Yuanyang Yin, Yaqi Zhao, Yajie Zhang, Ke Lin, Jiahao Wang, Xin Tao, Pengfei Wan, Di Zhang, Baoqun Yin, and Wentao Zhang.
\newblock Sea: Supervised embedding alignment for token-level visual-textual integration in mllms.
\newblock \emph{arXiv preprint arXiv:2408.11813}, 2024.

\bibitem[Yu et~al.(2019)Yu, Xu, Yu, Yu, Zhao, Zhuang, and Tao]{yu2019activitynet}
Zhou Yu, Dejing Xu, Jun Yu, Ting Yu, Zhou Zhao, Yueting Zhuang, and Dacheng Tao.
\newblock Activitynet-qa: A dataset for understanding complex web videos via question answering.
\newblock In \emph{Proceedings of the AAAI Conference on Artificial Intelligence}, pages 9127--9134, 2019.

\bibitem[Yue et~al.(2023)Yue, Qu, Zhang, Fu, Huang, Sun, Su, and Chen]{yue2023mammoth}
Xiang Yue, Xingwei Qu, Ge Zhang, Yao Fu, Wenhao Huang, Huan Sun, Yu Su, and Wenhu Chen.
\newblock Mammoth: Building math generalist models through hybrid instruction tuning.
\newblock \emph{arXiv preprint arXiv:2309.05653}, 2023.

\bibitem[Yue et~al.(2024)Yue, Ni, Zhang, Zheng, Liu, Zhang, Stevens, Jiang, Ren, Sun, et~al.]{yue2024mmmu}
Xiang Yue, Yuansheng Ni, Kai Zhang, Tianyu Zheng, Ruoqi Liu, Ge Zhang, Samuel Stevens, Dongfu Jiang, Weiming Ren, Yuxuan Sun, et~al.
\newblock Mmmu: A massive multi-discipline multimodal understanding and reasoning benchmark for expert agi.
\newblock In \emph{Proceedings of the IEEE/CVF Conference on Computer Vision and Pattern Recognition}, pages 9556--9567, 2024.

\bibitem[Zhang et~al.(2021)Zhang, Li, Hu, Yang, Zhang, Wang, Choi, and Gao]{zhang2021vinvl}
Pengchuan Zhang, Xiujun Li, Xiaowei Hu, Jianwei Yang, Lei Zhang, Lijuan Wang, Yejin Choi, and Jianfeng Gao.
\newblock Vinvl: Revisiting visual representations in vision-language models.
\newblock In \emph{Proceedings of the IEEE/CVF conference on computer vision and pattern recognition}, pages 5579--5588, 2021.

\bibitem[Zhang et~al.(2024)Zhang, Dong, Zang, Cao, Qian, Chen, Guo, Duan, Wang, Ouyang, et~al.]{zhang2024internlm}
Pan Zhang, Xiaoyi Dong, Yuhang Zang, Yuhang Cao, Rui Qian, Lin Chen, Qipeng Guo, Haodong Duan, Bin Wang, Linke Ouyang, et~al.
\newblock Internlm-xcomposer-2.5: A versatile large vision language model supporting long-contextual input and output.
\newblock \emph{arXiv preprint arXiv:2407.03320}, 2024.

\bibitem[Zhou et~al.(2018)Zhou, Xu, and Corso]{zhou2018towards}
Luowei Zhou, Chenliang Xu, and Jason Corso.
\newblock Towards automatic learning of procedures from web instructional videos.
\newblock In \emph{Proceedings of the AAAI Conference on Artificial Intelligence}, 2018.

\bibitem[Zhou et~al.(2025)Zhou, Zhu, Zhu, Wen, Liu, Xu, Meng, Cheng, Peng, Shen, and Feng]{zhou2025chatvlaunifiedmultimodalunderstanding}
Zhongyi Zhou, Yichen Zhu, Minjie Zhu, Junjie Wen, Ning Liu, Zhiyuan Xu, Weibin Meng, Ran Cheng, Yaxin Peng, Chaomin Shen, and Feifei Feng.
\newblock Chatvla: Unified multimodal understanding and robot control with vision-language-action model, 2025.

\end{thebibliography}
